\documentclass[runningheads]{llncs}

\usepackage{eccv}

\usepackage{eccvabbrv}

\usepackage{tablefootnote}
\usepackage{helvet}
\usepackage{courier}
\usepackage{amsmath}

\usepackage{amsthm}
\usepackage{amssymb}
\usepackage{mathrsfs}
\usepackage{color}
\usepackage{dsfont}
\usepackage{algorithmic}
\usepackage{algorithm}
\usepackage{xcolor}
\usepackage{amsmath}
\usepackage{makecell}
\usepackage{graphicx}
\usepackage{float}
\usepackage{bbm}
\usepackage{multirow}
\usepackage{booktabs}
\usepackage{marvosym}
\usepackage{ifsym}

\usepackage[accsupp]{axessibility}  

\usepackage{hyperref}

\usepackage{orcidlink}
\newcommand{\sys}{FairViT }

\begin{document}

\title{FairViT: Fair Vision Transformer via Adaptive Masking} 

\titlerunning{FairViT: Fair Vision Transformer via Adaptive Masking}

\author{Bowei Tian\inst{1}\orcidlink{0009-0005-7275-7955} \and
Ruijie Du\inst{2}\orcidlink{0009-0004-9451-0542} \and
Yanning Shen\inst{2}$^{(\text{\Letter})}$\orcidlink{0000-0002-7333-893X}}

\authorrunning{B.~Tian et al.}

\institute{Wuhan University, Wuhan, Hubei 430072, China \\
\email{boweitian@whu.edu.cn}\and
University of California, Irvine, CA 92697, USA \\
\email{\{ruijied,yannings\}@uci.edu}\\
}

\maketitle

\begin{abstract}
  Vision Transformer (ViT) has achieved excellent performance and demonstrated its promising potential in various computer vision tasks. The wide deployment of ViT in real-world tasks requires a thorough understanding of the societal impact of the model. However, most ViT-based works do not take fairness into account and it is unclear whether directly applying CNN-oriented debiased algorithm to ViT is feasible. Moreover, previous works typically sacrifice accuracy for fairness. Therefore, we aim to develop an algorithm that improves accuracy without sacrificing fairness. In this paper, we propose FairViT, a novel accurate and fair ViT framework. To this end, we introduce a novel distance loss and deploy adaptive fairness-aware masks on attention layers updating with model parameters. Experimental results show \sys can achieve accuracy better than other alternatives, even with competitive computational efficiency. Furthermore, \sys achieves appreciable fairness results.
  \keywords{Vision Transformer \and Accuracy \and Fairness \and Adaptive Masking}
\end{abstract}

\section{Introduction}

Vision transformer (ViT) \cite{dosovitskiy2020image,liu2021swin} has been widely adopted in various computer vision (CV) tasks, and is considered a viable alternative to the Convolutional Neural Network (CNN) \cite{he2016deep}. Unlike CNN, ViT has a specialized structure that can extract global relationships via a self-attention mechanism, leading to improved performance in various CV tasks, including image classification \cite{liu2021swin,moayeri2022comprehensive,bhojanapalli2021understanding}, object detection \cite{dai2021up,beal2020toward, fang2021you} and instance segmentation \cite{strudel2021segmenter,wang2021end,gu2022multi}. 
Due to its excellent performance, the structure has formed the architectural backbone of many CV algorithms for real-world applications. However, wide deployment of CV algorithms highly depends on how trustworthy they are \cite{qiang2023fairness,sudhakar2023mitigating}.  This prompts an investigation of the fairness aspects on ViT models.


Despite the abundance of debiased algorithms targeting Convolutional Neural Networks (CNNs) \cite{wang2022fairness,park2022fair,chen2022debiased}, there is a lack of literature concerning debiased algorithms on vision transformers.
Different from CNNs that capture pixel-wise local features through convolutions, vision transformers extract global contextual information through image patches. Vision transformers interpolate these patches through the attention mechanism with a stronger shape recognition capacity \cite{xie2022vit}. It is unclear whether directly applying CNN-oriented debiased algorithm to vision transformers is feasible \cite{chen2022debiased}.  Besides, vision transformers are shown to be more robust to input perturbations and latent features than CNNs \cite{dosovitskiy2020image,shao2021adversarial}, which may be challenging for a specific fair ViT design.

Fairness in ViT is investigated in several recent works \cite{sudhakar2023mitigating,roh2023dr,qiang2023fairness}, yet a majority of them either sacrifice accuracy for fairness, or require a huge amount of computational cost. 
TADeT, a targeted alignment technique proposed in \cite{sudhakar2023mitigating}, seeks to identify and eliminate bias from the query matrix in ViT. Their results demonstrate an effective debiased performance, and it is easy to implement in real scenarios. Whereas their directly manipulating parameters in the model will sacrifice accuracy for fairness. 
A bilevel optimization is designed in \cite{roh2023dr}, which finds the optimal data sampling ratios between real and generated data and leads to an improving tradeoff between fairness and accuracy, yet this method require relatively high computing power. 
Debiased Self-Attention (DSA) proposed in \cite{qiang2023fairness} is a fairness-target approach that enforces ViT to eliminate spurious features correlated with the sensitive label. DSA uses adversarial machine learning to enhance fairness-accuracy balance. However, it requires costly two-stage training, which is hard to deploy in real scenarios.
 
To address the aforementioned challenges, we propose FairViT, including adaptive masking and distance loss, are innovative and effective frameworks aimed at addressing fairness and accuracy concerns. Rather than deploying a high computing mechanism, the distance loss is a regularizer that is convenient to deploy, and the adaptive masking is convenient to calculate as well. With the assistance of adaptive masking, the model can reach a better performance on fairness and accuracy matrices. At the same time, distance loss is an extendable, convenient approach that can be used in other applications, not limited in ViT. The code is available at \href{https://github.com/abdd68/Fair-Vision-Transformer}{https://github.com/abdd68/Fair-Vision-Transformer}.

Our main contributions can be summarized as follows:
\begin{itemize} 
\item[$\bullet$] We introduce an adaptive masking framework wherein group-specific masks and weights are learned to enhance fairness. We equip the adaptive masking with a backward algorithm that optimizes the masks and weights.
\item[$\bullet$] We incorporate an extendable distance loss function manipulating the output scores to augment accuracy. 
\item[$\bullet$] We conduct extensive experiments on real datasets and demonstrate \sys achieves accuracy better than alternatives, even with competitive computational efficiency. Furthermore, \sys attains appreciable fairness results. 
\end{itemize}
\section{Related Work}
\subsection{Vision Transformer}
The transformer architecture was initially designed for natural language processing (NLP) \cite{vaswani2017attention} tasks. 
Unlike convolutional neural networks, the transformer network relies on the attention mechanism to process sequences of input tokens in parallel. 

Recently, the transformer architecture has been adapted to computer vision tasks \cite{dosovitskiy2020image}, utilizing the self-attention mechanism to model relationships between different parts of an image. ViT’s advantages include flexibility in handling various resolutions, capturing global information, parameter efficiency, and potential for better generalization. In many scenarios, ViT outperforms CNN and achieves considerable robustness \cite{raghu2021vision}. 

In ViT, sample $\mathbf{x}$ contains $p$ input image patches. ViT first employs an embedding layer to each patch to convert it into a embedding vector. Subsequently, ViT applies a series of transformer encoder layers to the embeddings, and each encoder layer consists of two parts: a Multi-Head Attention mechanism (MHA) and a position-wise FeedForward Network (FFN). The MHA layer models the interactions between the patch embeddings using self-attention, while the FFN layer implements a non-linear transformation on each patch embedding individually. The self-attention mechanism \cite{dosovitskiy2020image} can be illustrated as: 
\begin{equation}
    \text{Attn}(\mathbf{x}) = \text{S} \left( \frac{Q K^T}{\sqrt d} \right) V,
\end{equation}
where $Q$, $K$, and $V$ are the query, key, and value matrices, respectively, $\text{S}(\cdot)$ is the softmax function, $d$ is the dimension of the key vectors. Self-attention is an important building block for transformers and raises a huge amount of interest in the CV domain, since it is shown that the reliance on CNNs is not necessary and a pure transformer applied directly to sequences of image patches can perform very well on image classification tasks \cite{dosovitskiy2020image}. There are abundant works that aim to explore the transformers' attention mechanism, such as Gradient Attention Rollout \cite{abnar2020quantifying} in explainability and Swin Transformer \cite{liu2021swin} in efficiency.

\subsection{Fairness in Neural Networks}
Most of the existing debiasing methods for image classification tasks are specified for CNN or deep neural network (DNN) models \cite{wang2022fairness,park2022fair}, and can not be directly applied to ViTs.
However, several studies show that CV models make predictions by mixing sensitive features with input features \cite{zhao2017men,park2022fair}, and the sensitive features may capture biased relationships between the input features and the target labels. For example, the sensitive feature ``gender'' usually influences the accuracy of a face recognition task. In this case, it may lead to discriminatory results towards underrepresented groups, which causes serious social and ethical problems.

Fairness of ViT has been investigated in several recent works \cite{sudhakar2023mitigating,roh2023dr,qiang2023fairness}. 
  A targeted alignment technique TADeT was proposed in  \cite{sudhakar2023mitigating}, which seeks to identify and eliminate bias from the query matrix in ViT. However, their directly manipulating $Q$ sacrifices accuracy for fairness. Dr-Fairness \cite{roh2023dr} proposes a bilevel optimization that finds the optimal data sampling ratios between real and generated data, and it leads to an improving tradeoff between fairness and accuracy, yet they require relatively high computing power. Debiased Self-Attention (DSA) \cite{qiang2023fairness} is a fairness-target approach that enforces ViT to eliminate spurious features correlated with the sensitive label, and DSA uses adversarial machine learning to enhance fairness-accuracy balance.

 In this paper, we present a novel fair and accurate training framework designed for vision transformers. \sys outperforms existing works by demonstrating superior accuracy and appreciable fairness. Furthermore, our time cost experiment and multi-task testing show that \sys is applicable in real deployments, and maintains reasonable computational efficiency.
\section{Problem formulation}
We formulate the fairness-accuracy issue as a supervised classification problem, where the goal is to train a model $f$ using training samples \{$\mathbf{x}, s, y$\} and learn patterns from the data in order to make predictions, where $\mathbf{x}$ is the input feature, $y$ is the target label, and $s$ is a sensitive label. Let $y$ belongs to the space $\mathbf{Y}$, and $s$ belongs to the space $\mathbf{S}$, some examples of $\mathbf{S}$ include gender, race, or other attributes that can determine a sensitive group. We assume that $s$ can only be accessed in the training phase, and are not accessible in the validation or testing phase. The classification framework in training embodies the following form: 
\begin{equation}
    \min_{\boldsymbol{\theta}} L(f(\mathbf{x};\boldsymbol{\theta}),s,y)
\end{equation}
where $f(\mathbf{x};\boldsymbol{\theta})$ is the learned model parameterized by $\boldsymbol{\theta}$,  $L$ is the loss function characterizing the discrepancy between the estimated label and the target label. One common selection for $L$ is the cross-entropy loss \cite{mao2023cross}. However, the cross-entropy loss does not take $s$ into account \cite{du2022fairness}. 
Therefore, our objective is to devise a novel framework $f(\mathbf{x};\boldsymbol{\theta})$ to alleviate bias, and we use $s$ in adaptive masking during the back propagation. 
During the validation and testing stage, since the sensitive attribute $s$ is not available, the model treats $s$ as $\varnothing$ and computes the weighted sum within the adaptive masking.

\section{Fairness-aware Vision Transformer Design}

\begin{figure}[t]
    \centering
    \scriptsize
	\includegraphics[trim=0mm 0mm 0mm 0mm, clip,width=1\linewidth]{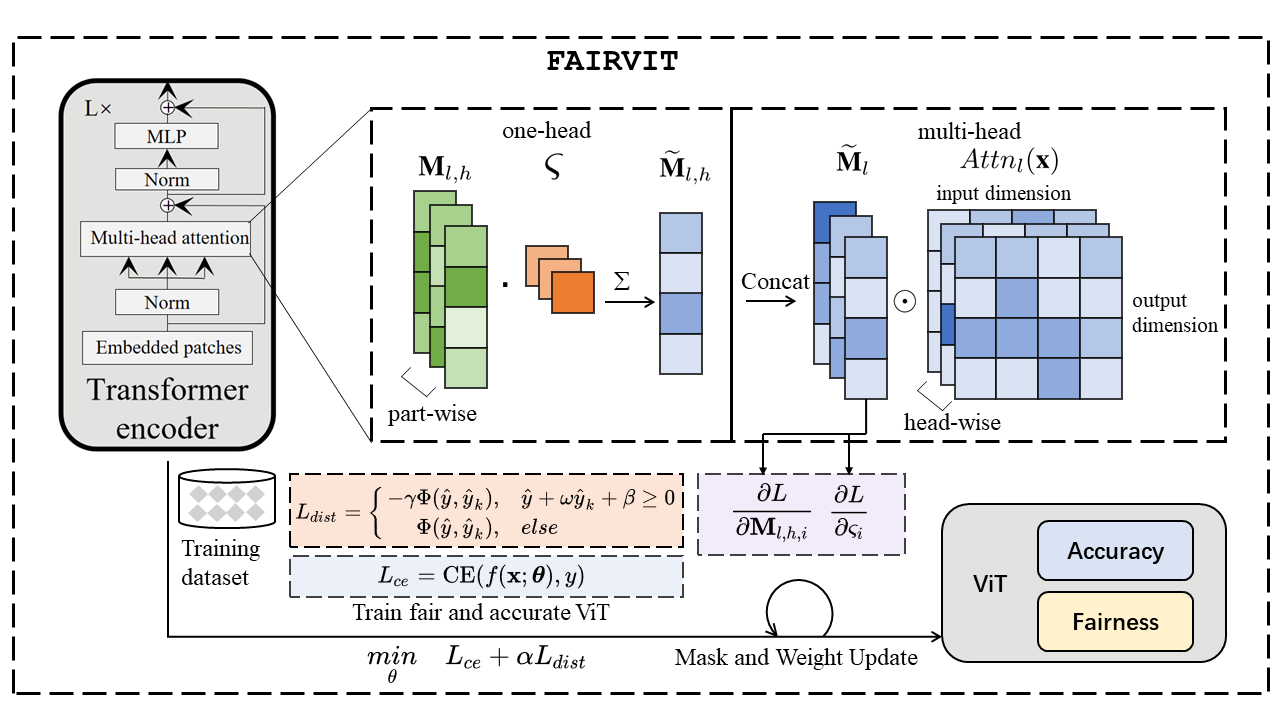}\\
	\caption{An illustration of \sys. For the forward propagation, we first apply weight $\varsigma$ to $\textbf{M}_{l,h}$, calculate the weighted sum $\widetilde{\textbf{M}}_{l,h}$, which is utilized to assist attention mechanism to control the information flow. For the backward propagation, we optimize $\mathbf{M}_{l,h,i}$ and $\varsigma_i$. Additionally, we introduce a novel distance loss $L_{dist}$.}
	\label{fig:overview}
\end{figure}

Our design comprises two pivotal parts, i.e. the adaptive masking and distance loss. First, we introduce adaptive masking, which is an assistance of the attention mechanism, concentrating on manipulating the model structure to enhance accuracy and maintain fairness. We optimize the adaptive masking by updating the masks and weights iteratively. Then, the distance loss is introduced to further enhance the accuracy. Figure~\ref{fig:overview} illustrates the overall procedure of FairViT. Algorithm~\ref{alg:trigger injection} outlines the entire fairness-aware process of FairViT.
\begin{algorithm}[tt]
	\caption{Pseudo-code of \sys} \label{alg:trigger injection}
	\begin{algorithmic}[1]
		\REQUIRE Transformer model parameters $\boldsymbol{\theta}$, training data set $T_t$, validation data set  $T_v$, 
        threshold $t$, epoch $E$ and learning rate $lr$.

		\ENSURE The accurate and fair model parameters $\boldsymbol{\theta}^*$.
        \STATE Initialize $\mathbf{M}_{l,h,i}=\mathbf{0}$, $L = INF$, $h=0$
        \WHILE {$h < E$ \textbf{and} $L > t$}
        \STATE // Training Stage
          \FORALL{$\mathbf{x} \in T_t$}
            \IF{h = 0}
            \STATE $L = L_{ce}$.
            \STATE Obtain $\frac{\partial L}{\partial \boldsymbol{\theta}}$ and back-propagate.
            \ELSE
            \STATE $L = L_{ce} + \alpha \cdot L_{dist}$.
            \STATE Obtain  $\frac{\partial L}{\partial \boldsymbol{\theta}}$, $\frac{\partial L}{\partial{\mathbf{M}_{l,h,i}}}$, $\frac{\partial L}{\partial{\varsigma_i}}$ by Equation (\ref{eq:7}-\ref{eq:8}) and back-propagate.
            \ENDIF
         \ENDFOR
         \STATE // Validation stage
         \FORALL{$\mathbf{x} \in T_v$} 
         \STATE $\hat{y} = f_{y}(\mathbf{x};\boldsymbol{\theta})$.
         \STATE $\hat{y}_k = \sum_{i\in\{topk\}/\{y\}} f_{i}(\mathbf{x};\boldsymbol{\theta})$.
         \ENDFOR
         \STATE Update $w$ and $\beta$ in Equation (\ref{eq:hyper}) by classifying $(\hat{y},\hat{y}_k) \rightarrow z$.
        \STATE $h = h+1$.
        \ENDWHILE
        \STATE $\boldsymbol{\theta}^* = \boldsymbol{\theta}$.
	    \RETURN $\boldsymbol{\theta}^*$.
  \end{algorithmic}
\end{algorithm}
\subsection{Adaptive Masking}
\label{sec:adaptivemasking}
\subsubsection{Forward Propagation Design}
Existing ViTs have demonstrated remarkable capabilities in various image recognition tasks. However, ViTs rely heavily on vast datasets for training, and its attention mechanism extracts information from every processed patches of an image, potentially perpetuating biases inherent in those datasets. Consequently, the accuracy of ViTs might vary across different groups, leading to disparities in performance. Therefore, we are seeking a solution to address this issue.
Motivated by multi-channel convolution \cite{vasudevan2017parallel}, where the convolution kernel channels align with the input channels, we seek to integrate analogous concepts into the ViT structure. This integration aims to enhance accuracy while upholding fairness. Our approach, named adaptive masking, first splits the training dataset into $G$ distinct parts, where each sensitive group has $\lfloor G/2 \rfloor$ parts. For each sensitive group, every part contains the same number of images. Since the number of samples in each sensitive group is different, the number of images in each part does not have to be equal between different sensitive groups. The splitting process is illustrated in Figure \ref{fig:data}. Subsequently, we associate each part with a corresponding mask and weight. Each part $i$ has a corresponding mask $\mathbf{M}_{l,h,i}$ and weight $\varsigma_i$ as parameters.
We then introduce one-Head Attention (HA) as the example:
\begin{figure}[t]
    \centering
    \includegraphics[width=0.7\linewidth]{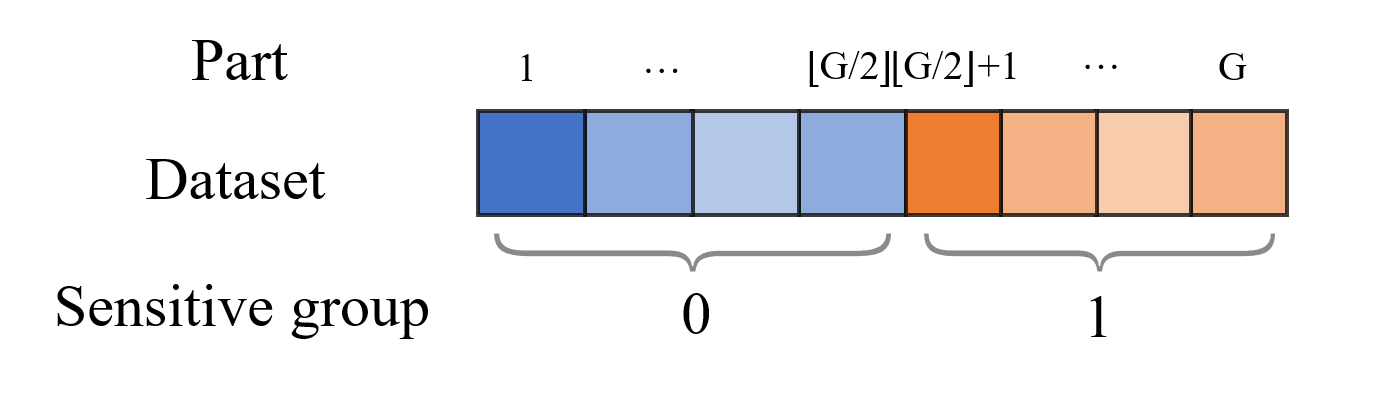}
    \caption{The split of the dataset in our design. Each part only contains samples from one sensitive group, and each part in one sensitive group contains the same number of images, but the number of images in one part between different sensitive groups does not have to be equal.\iffalse\shen{Group index starts from 0 or 1, 0 here, but start from 1 in (4)}\fi}
    \label{fig:data}
\end{figure}
\begin{equation}
\label{eq:attn}
    \text{Attn}_{l,h}(\mathbf{x}) = \text{S} \left( \frac{Q K^T}{\sqrt d} \right) V
\end{equation}
\begin{equation}
    \widetilde{\mathbf{M}}_{l,h} = {\sum_{i=1}^{G}(\varsigma_i \mathbf{M}_{l,h,i})} \label{eq:mm} 
\end{equation}
\begin{equation}
    \text{HA}(\mathbf{x}, \mathbf{M}_{l,h}) = \widetilde{\mathbf{M}}_{l,h}\odot \text{Attn}_{l,h}(\mathbf{x})
\end{equation}
where $\odot$ is the element-wise product, $\mathbf{M}_{l,h,i}$ represents the $i_{th}$ mask ($i \in \{1,\dots, G\}$) within layer $l$ and head $h$, $\varsigma_i$ is the weight of $\mathbf{M}_{l,h,i}$, and $\widetilde{\mathbf{M}}_{l,h}$ is the weighted sum of $\mathbf{M}_{l,h,i}$. We follow the Multi-Head Attention (MHA) as \cite{dosovitskiy2020image}, formulated as:
\begin{equation}
\begin{aligned}
    \text{MHA}  (\mathbf{x}, \mathbf{M}_{l}) = \Lambda(\text{HA}(\mathbf{x}, \mathbf{M}_{l,1}),\dots,\text{HA}(\mathbf{x}, \mathbf{M}_{l,H})) 
\end{aligned}
\end{equation}
where $\Lambda$ denotes the concatenate operation, $\text{HA}(\mathbf{x}, \mathbf{M}_{l,h}) \in \mathbb{R}^{p \times d}, \text{MHA}(\mathbf{x}, \mathbf{M}_{l}) \in \mathbb{R}^{p \times (Hd)}$, and $H$ is the number of heads in a transformer encoder layer. Adaptive masking can regulate information flow during forward propagation: if a particular group exhibits lower accuracy, the model trainer can potentially adapt by adjusting the weight assigned to this group. This adjustment framework establishes a criterion wherein adequate information is acquired for effective group classification, while simultaneously ensuring a fair balance of attention across the groups. Our subsequent objective is to guarantee that each mask and weight maintain an applicable distribution to attain the global optimum. Nonetheless, we observe that static values of $\varsigma$ demonstrate a sub-optimal performance in specific scenarios, as shown in Table \ref{tab:abl}. This observation motivates us to develop a gradient-based method that automatically optimizes the mask and weight.

\subsubsection{Updating the Adaptive Masking}
Instead of manually setting static masks and weights, we propose to update them iteratively during training. Specifically, 
given a sample belonging to part $g \in \{1,\dots,G\}$, the $\mathbf{M}_{l,h}$ and $\varsigma$ are updated through gradient descent. The gradient of $\mathbf{M}_{l,h,i}$ can be obtained as:
\begin{equation} 
\frac{\partial{L}}{\partial{\mathbf{M}_{l,h,i}}}=\left\{
\begin{aligned}
 \frac{\partial{L}}{\partial{\text{HA}}}&\text{Attn}_{l,h}(\mathbf{x}) \cdot \varsigma_i, & \text{if} \quad i = g \\
 &\textbf{0},\quad &\text{otherwise,} \\
\end{aligned}
\right.
\label{eq:7}
\end{equation}

\begin{figure}[t]
    \centering
    \includegraphics[width=0.6\linewidth]{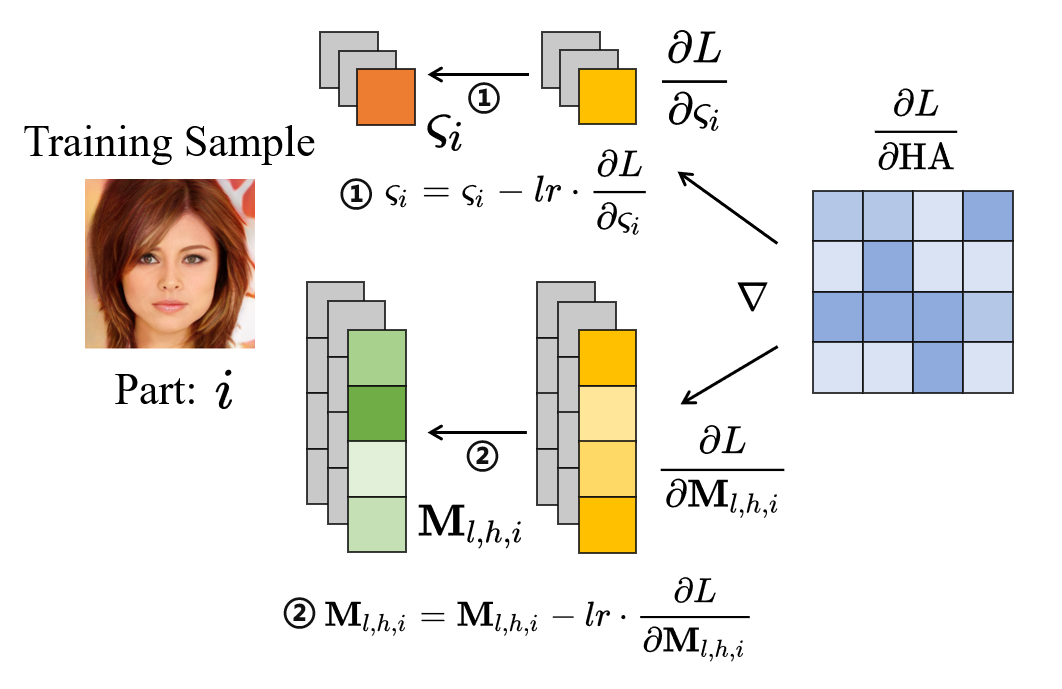}
    \caption{An illustration of the update process. We ascertain the specific part $i$ to which the training sample belongs, and $\nabla$ refers to the gradient calculation, specified in Equation (\ref{eq:7}-\ref{eq:8}). 
    The gray blocks signify that the gradients are zero during the backward pass of this training sample. 
    }
    \label{fig:back}
\end{figure}

To update $\varsigma_i$, we first obtain the computing map of $\varsigma_i$ towards $\widetilde{\mathbf{M}}_{l,h}$. Based on Equation \ref{eq:mm}, the gradient of $\varsigma_i$ can be obtained as
\begin{equation} 
\frac{\partial{L}}{\partial{\varsigma_i}}=\left\{
\begin{aligned}
 \sum_p\sum_d(\frac{\partial{L}}{\partial{\text{HA}}} &\text{Attn}_{l,h}(\mathbf{x}) \cdot (\sum_d\mathbf{M}_{l,h,i})), & \text{if} \quad i = g \\
 &0,\quad &\text{otherwise,} \\
\end{aligned}
\right.
\label{eq:8}
\end{equation}
An illustration of this process is shown as Figure \ref{fig:back}. Our method maintains reasonable computational efficiency, as $\frac{\partial{L}}{\partial{\text{HA}}}$ has been calculated during the backward propagation, requiring limited matrix multiplications to compute the gradients of $\mathbf{M}_{l,h,i}$ and $\varsigma_i$. 
Moreover, the experimental results in Table \ref{tab:abl} show that our iteratively updating the masks significantly improves the accuracy and upholds the fairness of the model compared with static masks. This improvement can be attributed to the parameterization of the $\mathbf{M}_{l,h}$ and $\varsigma$, i.e. they can be viewed as trainable parameters of the model, creating generalization to the sensitive groups.

\subsection{Distance Loss}
\label{sec:dist}
Cross-entropy loss uses the sign function that activates the output from the target label, deactivates the outputs from other labels. Solely minimizing cross-entropy loss actually omits the information from other labels. Therefore, we design the distance loss, considering not only maximizing the target label's score but also minimizing other labels' scores. Following some regularization techniques \cite{zafar2017fairness,du2022fairness}, we formulate a regularizer that improves accuracy. During the validation phase, we use logistic regression, a binary classifier to extract a hyperplane that underlines data distribution. Subsequently, in the training stage, we leverage this hyperplane to guide the processing of each sample. This innovative strategy allows us to align the training process more closely with the actual distribution of the data, enhancing adaptability and performance. 

In details, we define $\hat{y} = f_{{y}}(\mathbf{x};\boldsymbol{\theta})$ as the predicted score corresponding to the target label $y$. Additionally, we denote $\hat{y}_k = \sum_{i\in\{topk\}/\{y\}} f_i(\mathbf{x};\boldsymbol{\theta})$ as the cumulative score derived from labels in the top $k$ set, excluding the target label. In the validation stage, we train a linear classifier utilizing logistic regression, i.e. $\zeta(\hat{y},\hat{y}_k ) = \mathcal{S}(\hat{y} + \omega \hat{y}_k + \beta)$, where $\mathcal{S}(x) = \frac{1}{1+e^{-x}}$ is the sigmoid function, $\omega$ and $\beta$ are trainable parameters. The samples are labeled by {$z= \mathbbm{1}(\hat{y} = {\max}_{i}(f_i(\mathbf{x};\boldsymbol{\theta}))) \in \{0,1\}$,} indicating if the sample is classified correctly. The decision boundary of this linear classifier can be shown as 
\begin{equation}
\label{eq:hyper}
    \hat{y} + \omega \hat{y}_k + \beta = 0.
\end{equation}
Since $\omega$ and $\beta$ are updated during the validation stage, they remain constant in the training stage. Then we introduce the distance term as \cite{ballantine1952distance} for the training stage
\begin{equation}
\label{eqa:d}
    \Phi(\hat{y},\hat{y}_k) = \frac{|\hat{y} + \omega \hat{y}_k + \beta|}{\sqrt{1 + \omega^2} }
\end{equation}
which measures the distance between point $(\hat{y},\hat{y}_k)$ and the hyperplane in Equation (\ref{eq:hyper}). We train $\zeta$ in the validation stage to obtain $\omega$ and $\beta$, which remain constant in the next training stage. These fixed values facilitate the computation of our distance loss. 
To elaborate, the distance loss is as follows:

\begin{equation} 
L_{dist}=\left\{
\begin{aligned}
-\gamma \Phi(\hat{y},\hat{y}_k)\quad & \text{if } \hat{y} + \omega\hat{y}_k + \beta \geq 0, \\
\Phi(\hat{y},\hat{y}_k)\quad & \text{otherwise}, \\
\end{aligned}
\right.
\label{eqa:dist}
\end{equation}
where $\gamma$ is a non-negative hyperparameter. Minimizing $L_{dist}$ makes the points $(\hat{y},\hat{y}_k)$ that $\hat{y} + \omega\hat{y}_k + \beta \geq 0$ retain, and the points that $\hat{y} + \omega\hat{y}_k + \beta < 0$ are moved to the decision boundary, as our objective is to shift all the points to $\hat{y} + \omega\hat{y}_k + \beta \geq 0$. The overall loss function consists of two parts, shown as
\begin{equation}
    L = L_{ce} + \alpha L_{dist},
\end{equation}
where $L_{ce} = \text{CE}(f(\textbf{x};\boldsymbol{\theta}), y)$, and it guides the initial training phase when the hyperplane lacks meaningful definition. During the first epoch, $\boldsymbol{\theta}$ undergoes training solely based on $L_{ce}$. In each subsequent of the epochs, we update and refine the hyperplane, training the model with $L=L_{ce} + \alpha L_{dist}$.

Introducing the distance loss contributes to accuracy enhancement. Recall the definitions of $\hat{y}$ and $\hat{y}_k$, since the model selects the highest score's label as the predicted label, a higher $\hat{y}$ and a lower $\hat{y}_k$ suggest a greater likelihood of correct classification. Because we determine $\omega$ by the logistic regression on $z$, $\omega$ should assume a negative value, resulting in a positive slope for the hyperplane in Equation (\ref{eq:hyper}). Equation (\ref{eqa:d}) is the distance between $(\hat{y},\hat{y}_k)$ and the hyperplane. In Equation (\ref{eqa:dist}), $L_{dist}$ encourages both points that satisfy $\hat{y} + \omega\hat{y}_k + \beta < 0$ and $\hat{y} + \omega\hat{y}_k + \beta \geq 0$ to move towards  $\hat{y} + \omega\hat{y}_k + \beta \geq 0$.  In both scenarios, the distance loss encourages an increase in $\hat{y}$ and a decrease in $\hat{y}_k$, ultimately improving accuracy. Furthermore, we can extend the distance loss directly to other models, such as deep neural networks (DNNs) and convolutional neural networks (CNNs), as it solely requires the information from the model's output.
\section{Experiments}
\subsection{Experimental Setup}
We conduct experiments in three distinct scenarios on the CelebA dataset \cite{liu2015faceattributes}, a large-scale dataset of facial attributes, containing over 200,000 images of celebrities' faces. We attached our code in the supplementary materials and we will open-source it upon publication. In standard deployments, we initalize $\varsigma$ as $\mathbf{2}$ and clamp the weight to the range $(\epsilon, 4-\epsilon)$, where $\epsilon$ is a small value, set as $1e-8$ in our experiment. We initialize $\mathbf{M}_{l,h}$ with $\mathbf{1/2G}$ and clamp it to the range $(-1, 1)$, therefore $\widetilde{\mathbf{M}}_{l,h}$ is initialized with $\mathbf{1}$. We set the training-validation split ratio as $0.9:0.1$, $\gamma=0.5$, $G=10$ and $\alpha = 0.01$ in our experiments. For measuring fairness, the evaluation metric is as follows: \\
\noindent\textbf{Balanced Accuracy (BA)} \cite{park2020readme} measures the performance of a classification model, particularly when dealing with imbalanced datasets. Specifically, The formulation is shown as
\begin{equation}
\text{BA} = \frac{1}{4}(\text{TPR}_{s=0} + \text{TNR}_{s=0} + \text{TPR}_{s=1} +\text{TNR}_{s=1}).
\end{equation} It takes into account the imbalance in the dataset by calculating the average accuracy of each sensitive group and the target label.

\noindent\textbf{Demographic Parity (DP)} \cite{dwork2012fairness} measures how algorithms make predictions or decisions fairly among different demographic groups, or how much algorithms introduce biases or unfairness based on individual characteristics such as race, gender, and age, particularly between the sensitive groups (s = 0 and s = 1). Formally, \begin{equation}
    \text{DP} = |P(\hat{y}=1|s=1) - P(\hat{y}=1|s=0)|,
\end{equation} where $P$ denotes the probability calculated over the test set. A smaller DP typically means fewer differences between various groups in the outcomes of the algorithm.

\noindent\textbf{Equalized Opportunity ($\text{EO}$)} \cite{hardt2016equality} is a simple and interpretable notion of nondiscrimination with respect to a specified protected attribute. Specifically, \begin{equation}
    \text{EO} = |P(\hat{y}=1|s=1,y=1) - P(\hat{y}=1|s=0,y=1)|. 
\end{equation} As two of the most popular fairness matrices, DP focuses on the probability of being assigned to \textit{positive prediction} among different sensitive groups; EO focuses on the \textit{true positive rate} among different sensitive groups.

\subsection{Comparison with Baselines}
We select five SOTA fairness-aware baselines to compare with our work, i.e., Vanilla \cite{dosovitskiy2020image}, TADeT-MMD \cite{sudhakar2023mitigating}, TADeT \cite{sudhakar2023mitigating}, FSCL \cite{park2022fair} and FSCL+ \cite{park2022fair}. As the source code of DSA \cite{qiang2023fairness} is not publicly available, we did not include DSA in our experiment. We implement Vanilla, FSCL and FSCL+ based on their published source codes,  TADeT-MMD, and TADeT using illustrations in the paper. 
Table \ref{tab:base} demonstrates that FairViT exhibits superior fairness performance alongside appreciable accuracy. In comparison to FSCL+, which is our main competitor, \sys achieves a significantly higher accuracy of at least 4.5\%. In terms of fairness metrics, \sys showcases excellent unbiased effects.
\begin{table*}
\caption{The performance of image classification on CelebA dataset \cite{liu2015faceattributes} with Vanilla \cite{dosovitskiy2020image}, TADeT-MMD \cite{sudhakar2023mitigating}, TADeT \cite{sudhakar2023mitigating}, FCSL \cite{park2022fair}, FSCL+ \cite{park2022fair} and FairViT. Shown is the mean of 3 independent runs. Highlighted is the best result.}
\label{tab:base}
\centering
\footnotesize
\scalebox{0.85}{
    \begin{tabular}{c|cccc|cccc|cccc}
      \toprule
      \multirow{2}*{method} & \multicolumn{4}{c}{$\mathbf{Y}$: Attraction, $\mathbf{S}$: Gender} & \multicolumn{4}{c}{$\mathbf{Y}$: Expression, $\mathbf{S}$: Gender} & \multicolumn{4}{c}{$\mathbf{Y}$: Attraction, $\mathbf{S}$: Hair color} \\
      & $\text{ACC}_{\%}$ & $\text{BA}_{\%}$ & $\text{EO}_{e-2}$ & $\text{DP}_{e-1}$ & $\text{ACC}_{\%}$ & $\text{BA}_{\%}$ & $\text{EO}_{e-2}$ & $\text{DP}_{e-1}$ & $\text{ACC}_{\%}$ & $\text{BA}_{\%}$ & $\text{EO}_{e-2}$ & $\text{DP}_{e-1}$ \\
      \midrule
      Vanilla & 74.01& 72.36& 14.43& 3.245& 88.42&88.85&4.91&1.489&76.48&74.55&3.61&1.896\\
      TADeT-MMD & 79.89&73.85&7.10&3.693&92.51&93.03&2.48&1.290&77.97&75.64&2.27&1.491\\
      TADeT &78.73&74.52&3.11&3.116&90.05&90.68&4.86&1.443&78.49&77.42&3.78&1.057\\
      FSCL&79.09&74.76&1.78&3.004&89.37&90.08&1.76&1.344&78.85&78.06&2.65&0.989\\
      FSCL+&77.26&73.42&\textbf{0.79}&\textbf{2.604}&88.83&89.02&\textbf{1.20}&1.263&78.02&77.37&\textbf{1.79}&0.834\\
      FairViT&\textbf{84.01}&\textbf{79.96}&1.15&2.837&\textbf{94.27}&\textbf{94.12}&1.52&\textbf{1.205}&\textbf{82.52}&\textbf{81.56}& 2.10&\textbf{0.701}\\
      \bottomrule
    \end{tabular}}
\end{table*}

\subsection{Ablation Study}
\textbf{Method ablation study:} We conducted an ablation study to evaluate the effectiveness of adaptive masking and the distance loss. The results are presented in Table \ref{tab:abl}. Here, $\Theta$ denotes the adaptive masking method without updating masks and weights, while $\Delta\Theta$ signifies the deployment of adaptive masking with updating masks and weights. 
The experiment results show that both $L_{dist}$ and $\Delta\Theta$ contribute to accuracy improvement, with $\Delta\Theta$ playing a much more crucial role in enhancing accuracy and fairness. 
Continuously updating adaptive masks performs much better in both accuracy and fairness than static masks. 
\begin{table*}
\caption{Ablation study of FairViT. Shown is the mean of 3 independent runs. Highlighted is the best result.}
\label{tab:abl}
\centering
\footnotesize
\scalebox{0.8}{
    \begin{tabular}{l|cccc|cccc|cccc}
      \toprule
      \multirow{2}*{method} & \multicolumn{4}{c}{$\mathbf{Y}$: Attraction, $\mathbf{S}$: Gender} & \multicolumn{4}{c}{$\mathbf{Y}$: Expression, $\mathbf{S}$: Gender} & \multicolumn{4}{c}{$\mathbf{Y}$: Attraction, $\mathbf{S}$: Hair color} \\
      & $\text{ACC}_{\%}$ & $\text{BA}_{\%}$ & $\text{EO}_{e-2}$ & $\text{DP}_{e-1}$ & $\text{ACC}_{\%}$ & $\text{BA}_{\%}$ & $\text{EO}_{e-2}$ & $\text{DP}_{e-1}$ & $\text{ACC}_{\%}$ & $\text{BA}_{\%}$ & $\text{EO}_{e-2}$ & $\text{DP}_{e-1}$ \\
      \midrule
      $L_{ce}$ & 74.01& 72.36& 14.43& 3.245& 88.42&88.85&4.91&1.489&76.48&74.55&3.61&1.886\\
      $L_{ce} + L_{dist}$ & 77.01&72.68&12.54&3.166&89.49&89.98&3.85&1.426&77.08&74.89&2.10&1.741\\
     $L_{ce} + L_{dist} + \Theta$ &79.96&74.06&8.99&3.929&92.04&92.77&6.86&1.484&80.10&78.79& 7.98&2.051\\
      $L_{ce} + L_{dist} + \Delta\Theta$ & \textbf{84.01}&\textbf{79.96}&\textbf{1.15}&\textbf{2.837}& \textbf{94.27}& \textbf{94.12}& \textbf{1.52}& \textbf{1.268}&\textbf{82.52}&\textbf{81.56}&\textbf{2.10}&\textbf{0.701}\\
      \bottomrule
    \end{tabular}}
\end{table*}

\textbf{Impact of $\alpha$ in the loss function:} In Table \ref{tab:alpha}, we observe that as $\alpha$ increases, the accuracy gradually improves while $\text{EO}$ and BA are not hurt at all.  we can empirically observe that $\alpha$ is even beneficial to fairness at around 0.1. When $\alpha$ reaches a certain threshold (e.g. between 0.1 and 1 in the second case from Table \ref{tab:alpha}), the model experiences a decrease in both accuracy and fairness. This phenomenon could be attributed to the model's need to strike a balance between the distance loss and the cross-entropy loss. An inappropriate $\alpha$ might disrupt the optimization process, keeping the model from focusing on the optimization objective. For subsequent experiments, we set $\alpha=0.01$.

\begin{table}[t]
\caption{Impact of $\alpha$, the weight of distance loss. Shown is the mean of 3 independent runs. Highlighted is the best result.}
\label{tab:alpha}
\centering
\footnotesize
    \scalebox{0.9}{\begin{tabular}{c|cccc|cccc|cccc} 
      \toprule
      \multirow{2}*{$\alpha$} & \multicolumn{4}{c}{$\mathbf{Y}$: Attraction, $\mathbf{S}$: Gender} & \multicolumn{4}{c}{$\mathbf{Y}$: Expression, $\mathbf{S}$: Gender} & \multicolumn{4}{c}{$\mathbf{Y}$: Attraction, $\mathbf{S}$: Hair color} \\ 
      & $\text{ACC}_{\%}$ & $\text{BA}_{\%}$ & $\text{EO}_{e-2}$ & $\text{DP}_{e-1}$ & $\text{ACC}_{\%}$ & $\text{BA}_{\%}$ & $\text{EO}_{e-2}$ & $\text{DP}_{e-1}$ & $\text{ACC}_{\%}$ & $\text{BA}_{\%}$ & $\text{EO}_{e-2}$ & $\text{DP}_{e-1}$ \\
      \midrule
      0 & 81.08&77.19&4.89&3.027& 91.49&89.67&3.43&1.373&80.25&77.97&5.25&0.788\\ 
      0.001 & 81.15&79.19&4.03&\textbf{2.989}&92.18&89.81&3.60&1.320&82.25&\textbf{80.07}&\textbf{3.92}&1.009\\ 
      0.01 & \textbf{82.51}&\textbf{79.41}&4.48&3.337&93.00&92.87&2.40&1.424&\textbf{82.49}&79.62&5.08&0.942\\ 
      0.1& 81.73&76.51&\textbf{3.33}&3.156&\textbf{93.54}&\textbf{93.70}&\textbf{1.23}&\textbf{1.204}&80.35&77.43&4.24&\textbf{0.589}\\ 
      1& 75.53&70.38&5.92&3.597&82.63&83.80&3.34&1.228&75.18&73.49&5.87&1.053\\ 
      \bottomrule
    \end{tabular}}
\end{table}

\textbf{Impact of $\gamma$ in Distance Loss:}
In Table \ref{tab:gamma}, we note that as the values of $\gamma$ increase, both $\text{EO}$ and BA typically exhibit an initial rise until reaching 0.5, after which they decline. Meanwhile, accuracy follows an increasing trend till $0.5$, then it begins to decrease. We can observe $\gamma$ positively influences the impact of the distance loss, benefiting accuracy. However, excessively high values of $\gamma$ might create an imbalance between the two losses, potentially leading to decreased accuracy. Given these observations, we use $\gamma=0.5$ as it represents an optimal balance between fairness and accuracy based on our analysis.

\begin{table}[t]
\caption{Impact of $\gamma$. Shown is the mean of 3 independent runs. Highlighted is the best result.}
\label{tab:gamma}
\centering
\footnotesize
    \scalebox{0.9}{\begin{tabular}{c|cccc|cccc|cccc}
      \toprule
      \multirow{2}*{$\gamma$} & \multicolumn{4}{c}{$\mathbf{Y}$: Attraction, $\mathbf{S}$: Gender} & \multicolumn{4}{c}{$\mathbf{Y}$: Expression, $\mathbf{S}$: Gender} & \multicolumn{4}{c}{$\mathbf{Y}$: Attraction, $\mathbf{S}$: Hair color} \\
      & $\text{ACC}_{\%}$ & $\text{BA}_{\%}$ & $\text{EO}_{e-2}$ & $\text{DP}_{e-1}$ & $\text{ACC}_{\%}$ & $\text{BA}_{\%}$ & $\text{EO}_{e-2}$ & $\text{DP}_{e-1}$ & $\text{ACC}_{\%}$ & $\text{BA}_{\%}$ & $\text{EO}_{e-2}$ & $\text{DP}_{e-1}$ \\
      \midrule
      0.1 & 81.65& 77.82& 5.11& 3.386& 92.17& 91.31&2.94&1.556& 81.59&78.83&3.67&0.807\\
      0.3 & \textbf{82.89}& 78.96& \textbf{2.29 }& 3.205& 92.58& 91.53&1.75&1.296& 82.17&80.21&3.51&0.742\\
      0.5 & 82.75&\textbf{79.12}&3.06&\textbf{2.685}& \textbf{93.89}& \textbf{94.12}& \textbf{1.52}& 1.205&\textbf{82.51}&\textbf{80.91}&\textbf{3.07}&\textbf{0.535}\\
      0.7 & 82.46&77.23& 3.22& 2.909& 93.05&93.86&1.57&\textbf{1.016}&80.67&77.69&4.59&0.627\\
      0.9 & 81.67& 75.85& 3.31& 3.237& 92.25&92.73&2.02&1.338&80.83&76.84&4.87&0.939\\
      \bottomrule
    \end{tabular}}
\end{table}
\begin{figure}[h!]
  \subfloat
  {\includegraphics[width=0.49\textwidth]{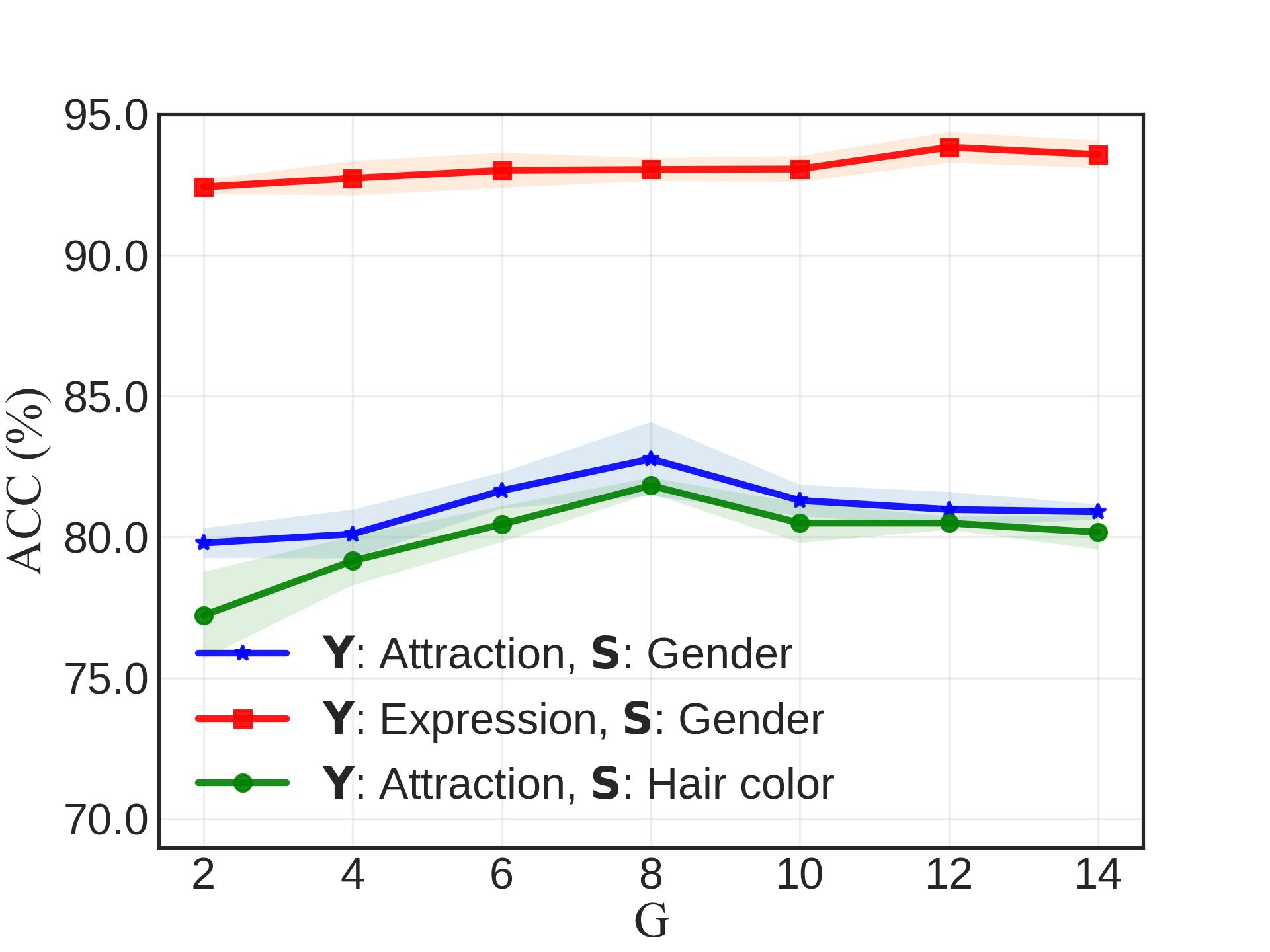}\label{fig:subfig7}}
  \subfloat
  {\includegraphics[width=0.49\textwidth]{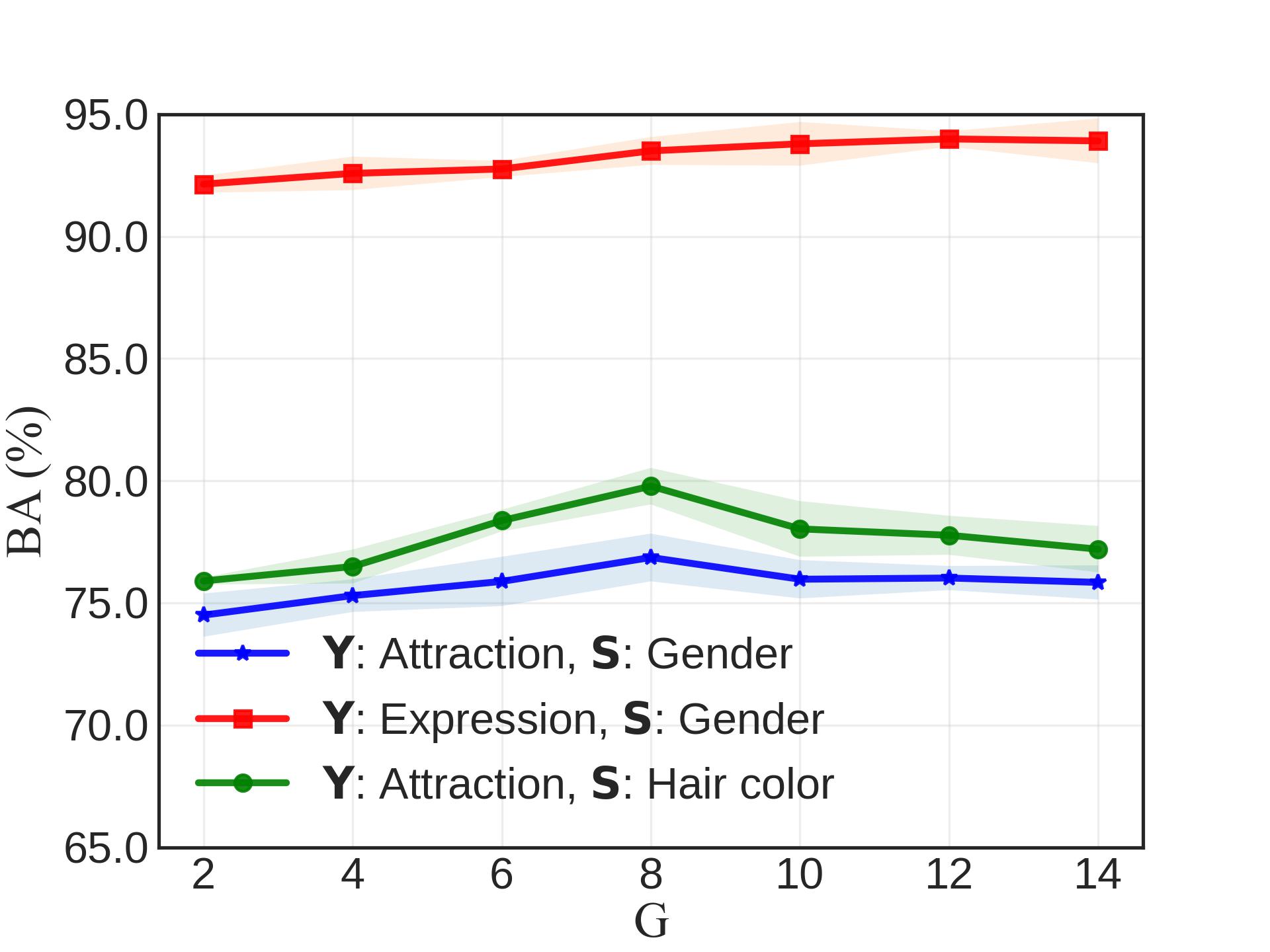}\label{fig:subfig8}}
\quad
  \subfloat
  {\includegraphics[width=0.49\textwidth]{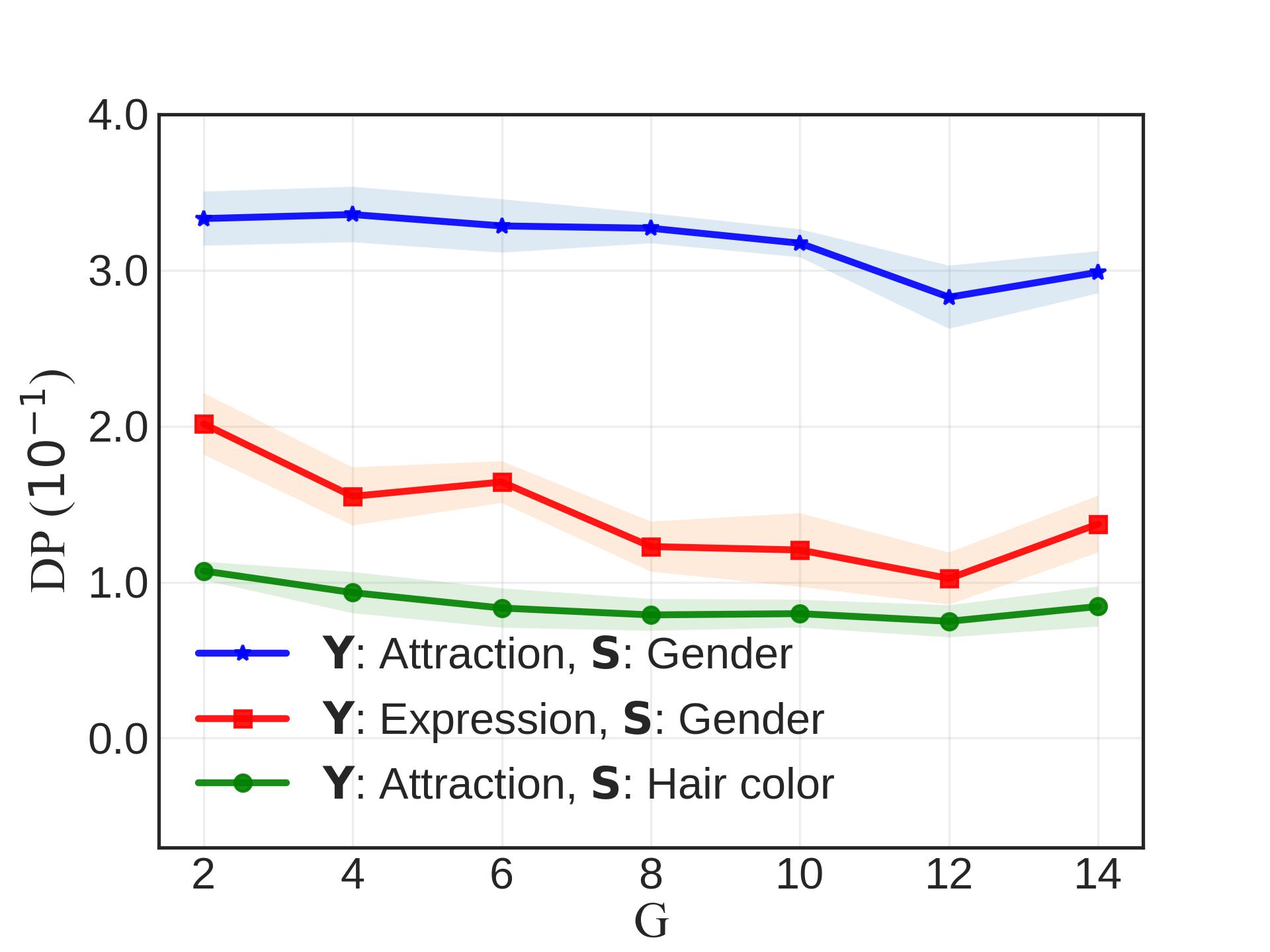}\label{fig:subfig9}}
  \subfloat
  {\includegraphics[width=0.49\textwidth]{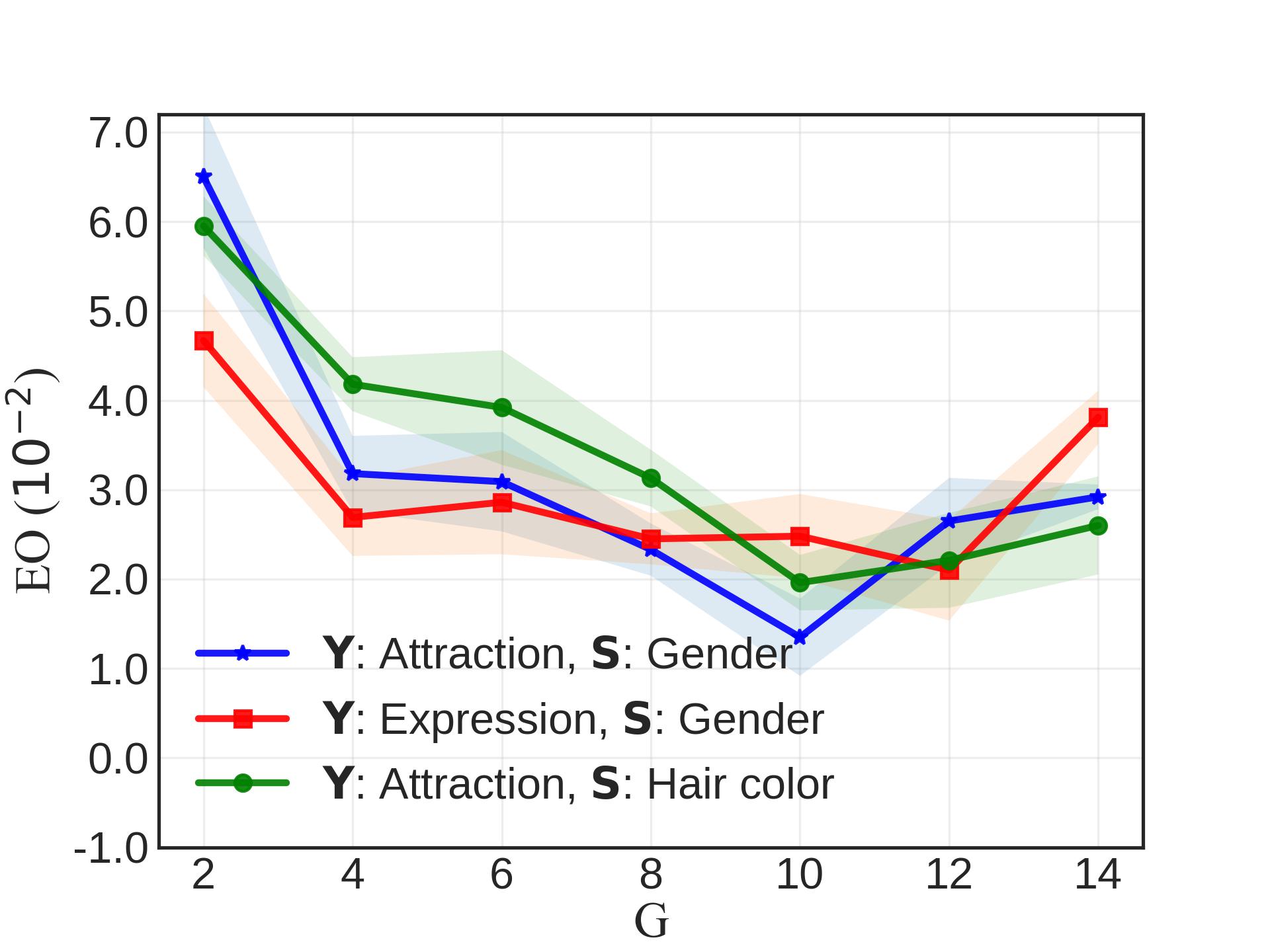}\label{fig:subfig10}}
  \caption{Impact of $G$. Shown is the mean $\pm$ standard deviation of 3 independent runs.}
  \label{tab:G}
\end{figure}
\textbf{Impact of $G$ in the adaptive masking:} In Figure \ref{tab:G}, we demonstrate the impact of $G$, the number of distinct parts, in the adaptive masking. When $G$ is small, the accuracy and fairness matrices do not bring huge benefits, however, as $G$ reaches a threshold, the accuracy and fairness tend to be stabilized and perform well. A possible explanation is that $G$ has a positive correlation with the adjustment capabilities of the model, as the model can consider more different parts at the same time and judge them in a more personalized way. However, an exceeding $G$ may result in too few images in one part, where each part will not have enough training data to obtain adequate performance, resulting in a slight decrease in performance. Therefore, the best choice of $G$ may vary across different problem scenarios and datasets. 
\begin{figure*}[t]
    \centering
    \scriptsize
	\includegraphics[trim=0mm 0mm 0mm 0mm, clip,width=1.03\linewidth]{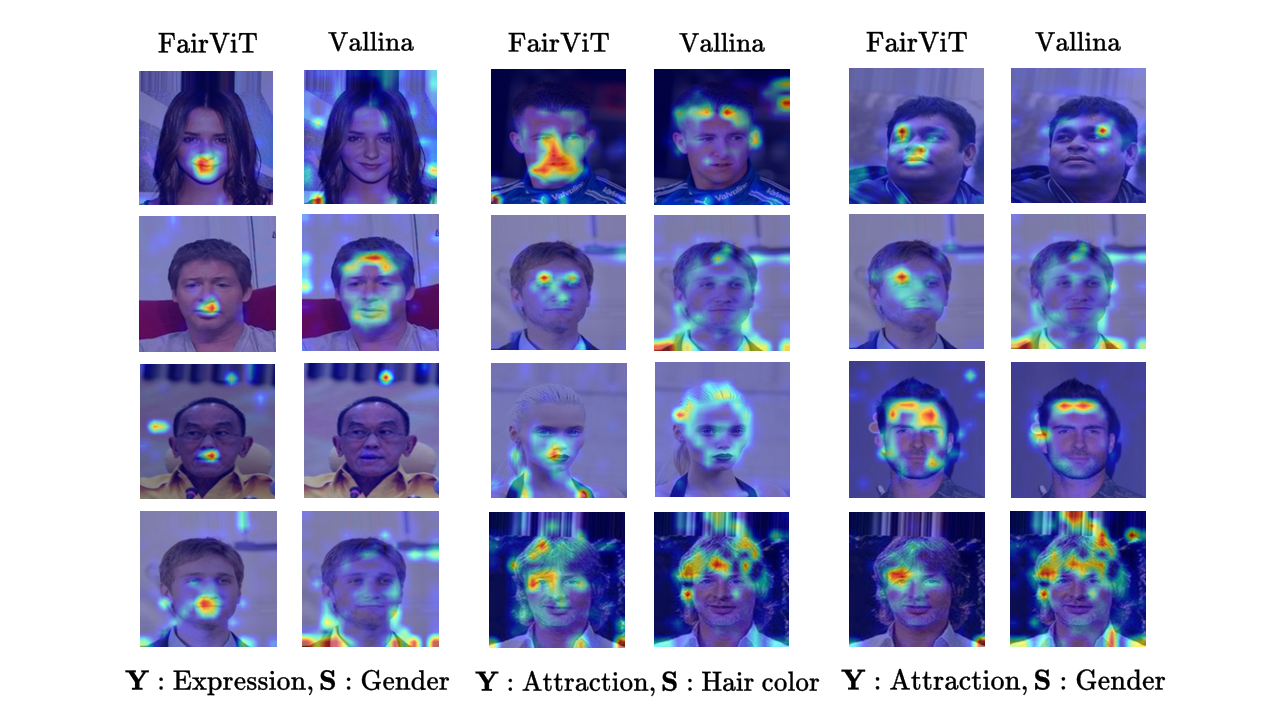}\\
	\caption{The interpretability study of \sys.}
	\label{fig:interpret}
\end{figure*}
\subsection{Interpretability Study}
\label{sec:interpret}
We conduct an interpretability study on \sys to elucidate the reasons behind its superior performance across diverse scenarios, shown in Figure \ref{fig:interpret}. We use Gradient Attention Rollout (GAR) \cite{abnar2020quantifying} to generate heat maps that accentuate crucial decision-making zones in ViT, and details about GAR are in Appendix A. We observe there are noteworthy distinctions of focused areas between Vanilla and FairViT. The Vanilla method appears to capture information relevant to sensitive attributes like $\text{Gender}$ in the first scenario and $\text{Hair color}$ in the second scenario. In contrast, FairViT, leveraging adaptive masking, demonstrates a tendency to extract information relevant to the target attributes, such as $\text{Expression}$ in the first scenario and $\text{Attraction}$ in the second scenario. This phenomenon illustrates the effectiveness of FairViT in fairness and accuracy. Furthermore, FairViT generates heat maps that are distributed more distinctly and densely in space, potentially indicating enhanced model learning.

\subsection{Time Cost}
To evaluate the efficiency of \sys, we conduct a comparative analysis of computational costs between \sys and the baselines, as illustrated in Table~\ref{tab:time}. Our findings reveal that \sys exhibits comparable computational costs to the baselines. Moreover, \sys attains a superior balance between accuracy and fairness while maintaining reasonable computational efficiency. Compared with FSCL+, \sys is 6 times faster while achieving better accuracy and competitive fairness results. The core incremental consumption in FairViT is the adaptive masking, requiring $O(p*d)$, where $p$ is the patch number and $d$ is the dimension of the key. The time complexity is better than FSCL, which requires cubic time complexity.
\begin{table}[h]
\caption{The time cost of  \sys compared with other baselines. Shown is the mean of 3 independent runs. Highlighted is the best result.}
\label{tab:time}
    \centering
    \scalebox{1}{\begin{tabular}{c|ccc}
    \toprule
        method & AG\tablefootnote{AG is short of $\mathbf{Y}$: Attraction, $\mathbf{S}$: Gender.}$_{\text{min}}$& EG\tablefootnote{EG is short of $\mathbf{Y}$: Expression, $\mathbf{S}$: Gender.}$_{\text{min}}$& AH\tablefootnote{AH is short of $\mathbf{Y}$: Attraction, $\mathbf{S}$: Hair color.}$_{\text{min}}$\\ 
        \midrule
        Vallina & \textbf{4.62}& \textbf{4.38}& \textbf{4.51}\\ 
        TADeT-MMD & 4.68& 4.50& 4.61\\ 
        TADeT & 7.74& 7.08& 7.38\\ 
        FSCL & 34.52& 34.84& 35.34\\ 
        FSCL+ & 34.73& 34.98& 35.79\\ 
        \sys &5.09&5.16&5.27\\
        \bottomrule
    \end{tabular}}
\end{table}
\section{Conclusion, Limitation and Discussion}
In this paper, we proposed FairViT, addressing the fairness-accuracy issue in vision transformers. FairViT employs adaptive masks to alleviate bias without compromising accuracy and crafts a versatile distance loss to enhance overall accuracy. Extensive experiments validate that FairViT can enhance fairness while upholding comparable levels of accuracy.

In the future, it would be interesting to extend the proposed techniques to a broader range of neural networks. In upcoming research endeavors, we aim to further explore the intrinsic mechanisms driving the effectiveness of distance loss and adaptive masking. Furthermore, more experimental evaluations on other learning tasks other than classification are also of great interest. For example, we plan to further explore fair generation tasks such as text-to-image generation\cite{friedrich2023fair} and graph generation\cite{kose2023fast, wang2023fairness}.
\clearpage

%
%


%
%
\bibliographystyle{splncs04}
\bibliography{main}

\begin{appendix}

\title{Supplementary Material for ``FairViT: Fair Vision Transformer via Adaptive Masking''} 

\titlerunning{Supplementary Material for FairViT}

\author{Bowei Tian\inst{1}\orcidlink{0009-0005-7275-7955} \and
Ruijie Du\inst{2}\orcidlink{0009-0004-9451-0542} \and
Yanning Shen\inst{2}$^{(\text{\Letter})}$\orcidlink{0000-0002-7333-893X}}

\authorrunning{B.~Tian et al.}

\institute{Wuhan University, Wuhan, Hubei 430072, China \\
\email{boweitian@whu.edu.cn}\and
University of California, Irvine, CA 92697, USA \\
\email{\{ruijied,yannings\}@uci.edu}\\
}

\maketitle

The supplemental material consists of this appendix. This appendix includes an illustration of interpretability study (Section \ref{sec:inter}), more experimental results on the ablation study of $\alpha$ and $\gamma$ (Section \ref{sec:abl}), and some implementation details (Section \ref{sec:implementation}).
\renewcommand\thefigure{\Alph{section}\arabic{figure}}
\renewcommand\theequation{\Alph{section}\arabic{equation}}
\section{Interpretability Study}
\label{sec:inter}
\subsection{Gradient Attention Rollout}
The Gradient Attention Rollout (GAR) \cite{bib1} aims to illustrate why the attention mechanism performs well in many scenarios in computer vision. GAR achieves interpretability by a heat map highlighting how much areas contribute to the output. Specifically, GAR is defined as
\begin{equation}
    \mathcal{A}_l=\begin{cases}
  \mathbf{A}_l(\mathbf{x})\frac{\partial \hat{y}}{ \partial \mathbf{A}_l(\mathbf{x})} \mathcal{A}_{l-1}, & \text{if } l > 0,
  \\
 \mathbf{A}_l(\mathbf{x})\frac{\partial \hat{y}}{ \partial \mathbf{A}_l(\mathbf{x})}, & \text{if }  l = 0,
  \end{cases}
\end{equation}
where $\mathbf{A}$ is an abbreviation of Attn in Equation (\ref{eq:attn}), and $\mathcal{A}_l$ denotes the GAR on the $l_{th}$ layer of the transformer.  To generate the heat map, we assign the value $\mathcal{A}_N^{0,i}$ to the $i_{th}$ patch in the image, where $\mathcal{A}_N$ represents the GAR of the last layer, measuring the importance of each patch in the final prediction. Note that $\mathcal{A}_N$ is a matrix, and $\mathcal{A}_N^{0,i}$ corresponds to the element at the 0-th row and $i$-th column. The primary objective of GAR is to quantify the relative importance of each input patch within the attention mechanism, and it is particularly useful for analyzing the model behavior and explaining its decision-making process \cite{bib1}.
\subsection{Additional Implementation}
\begin{figure*}[tbp]
    \centering
    \scriptsize
	\includegraphics[trim=0mm 0mm 0mm 0mm, clip,width=1.03\linewidth]{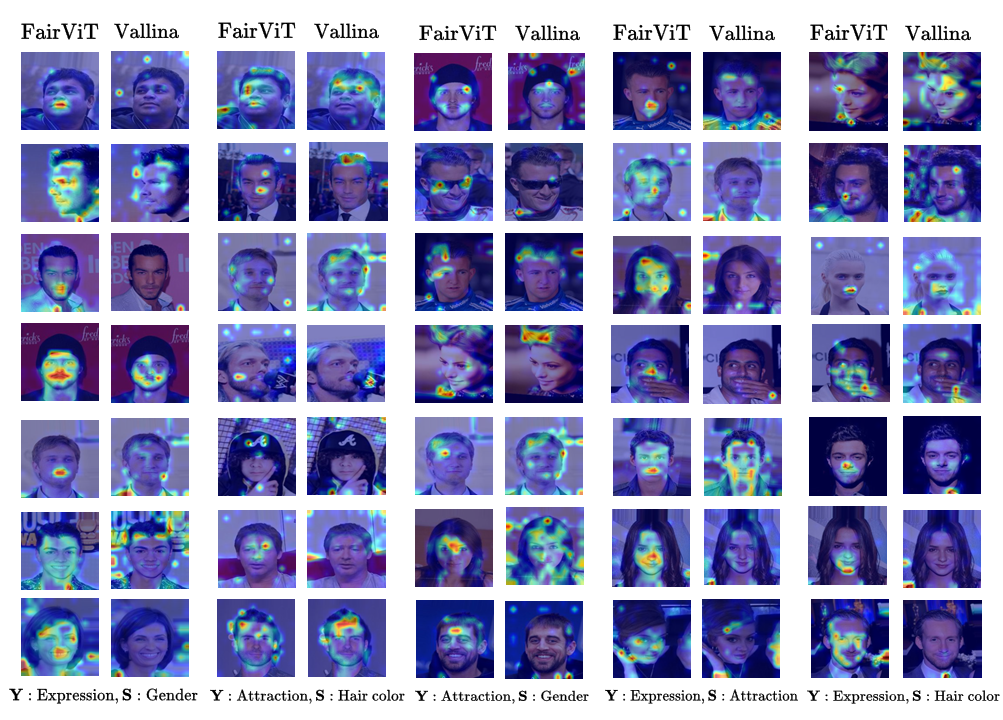}\\
	\caption{The extended interpretability study of \sys.}
	\label{fig:interpret2}
\end{figure*}

In this section, we implement the interpretability study into two additional scenarios, i.e. $\mathbf{Y}$: Expression, $\mathbf{S}$: Attraction and $\mathbf{Y}$: Expression, $\mathbf{S}$: Hair color, to evaluate the effectiveness of FairViT. The corresponding images are shown in Figure \ref{fig:interpret2}, and the outcomes align consistently with our observations in Section \ref{sec:interpret}. The Vanilla method captures information relevant to sensitive attributes. In contrast, FairViT exhibits a tendency to extract information relevant to the target attributes, such as $\text{Expression}$ in the first scenario and $\text{Attraction}$ in the second scenario. Furthermore, from the last two scenarios, despite the variations in the sensitive attribute, the heat map remains capable in capturing the target attribute.
\section{Ablation Study: Standard Deviation Qualification}
\label{sec:abl}
\begin{figure}[t]
  \subfloat
  {\includegraphics[width=0.49\textwidth]{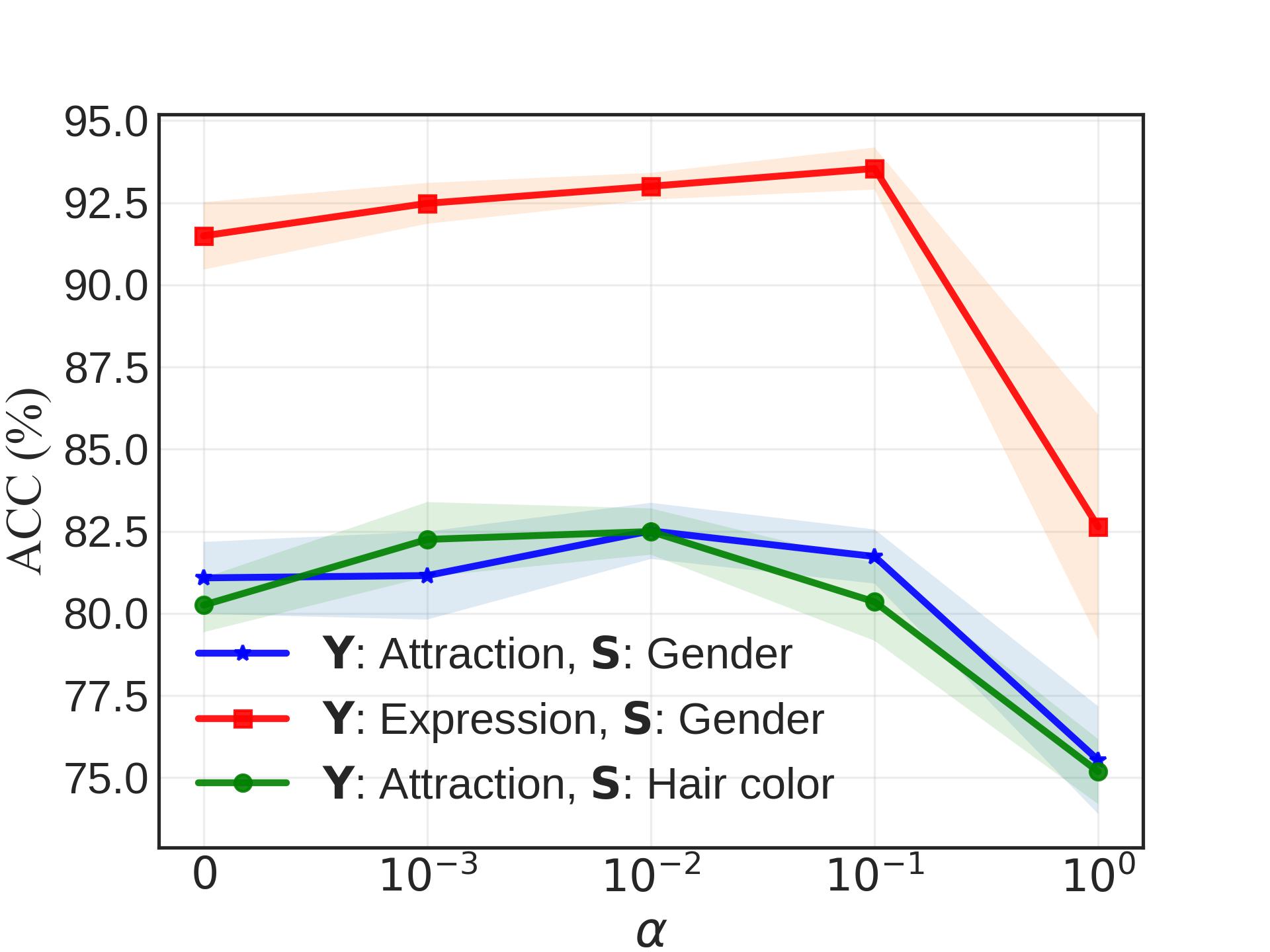}\label{fig:subfig7}}
  \subfloat
  {\includegraphics[width=0.49\textwidth]{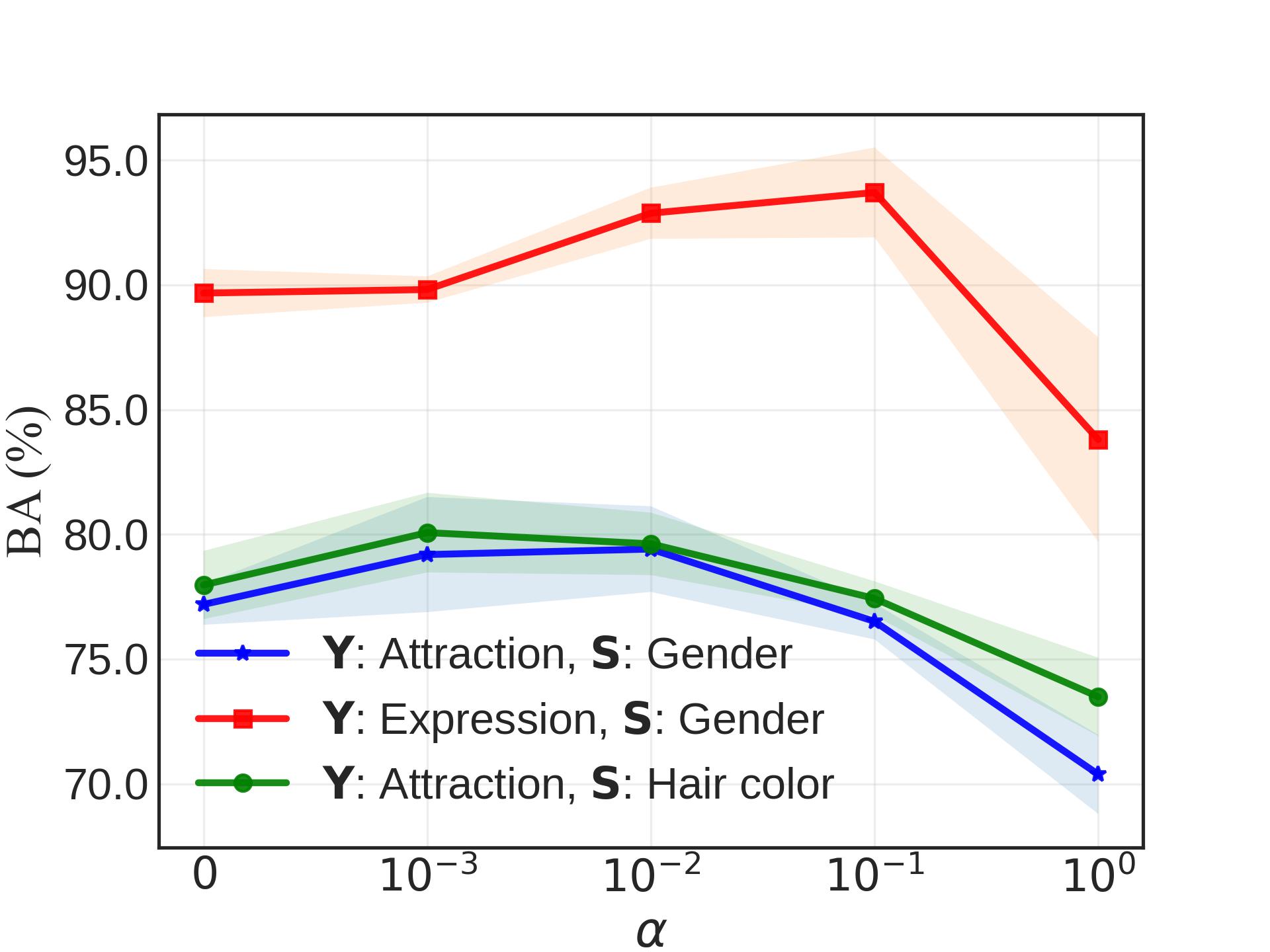}\label{fig:subfig8}}
\quad
  \subfloat
  {\includegraphics[width=0.49\textwidth]{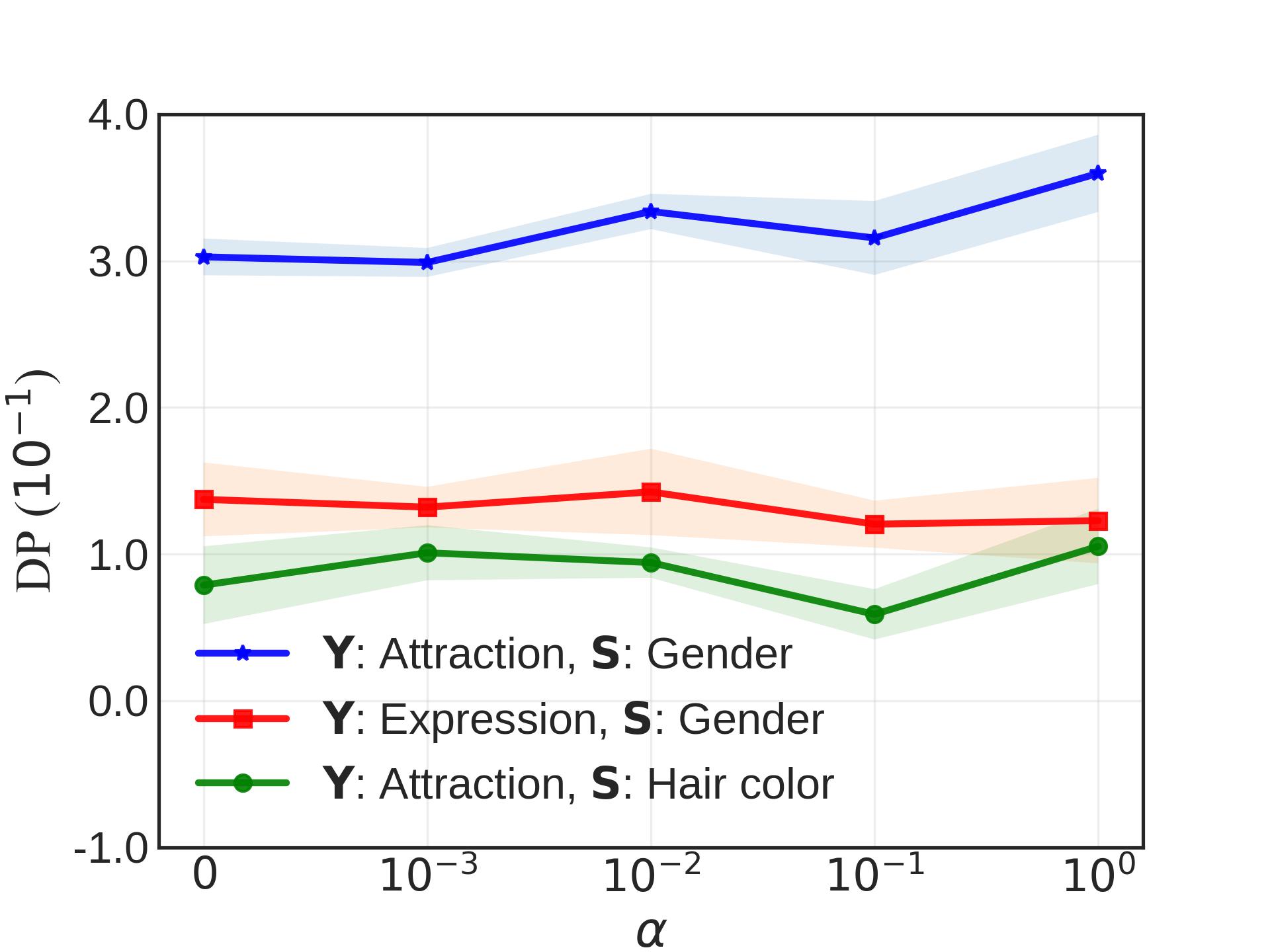}\label{fig:subfig9}}
  \subfloat
  {\includegraphics[width=0.49\textwidth]{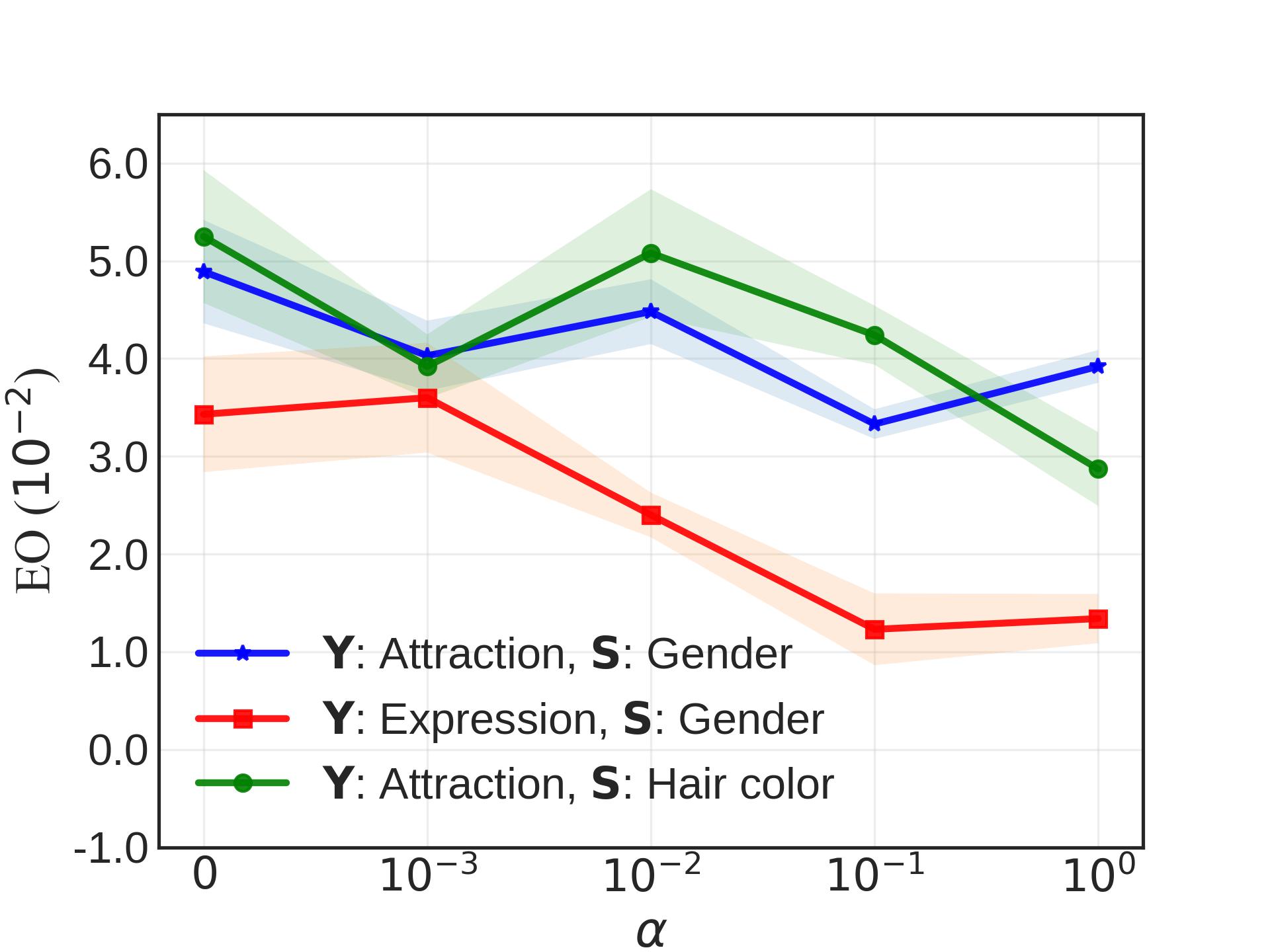}\label{fig:subfig10}}
  \caption{Impact of $\alpha$. Shown is the mean $\pm$ standard deviation of 3 runs with different random seeds.}
  \label{fig:alpha}
\end{figure}

\begin{figure}[t]
  \subfloat
  {\includegraphics[width=0.49\textwidth]{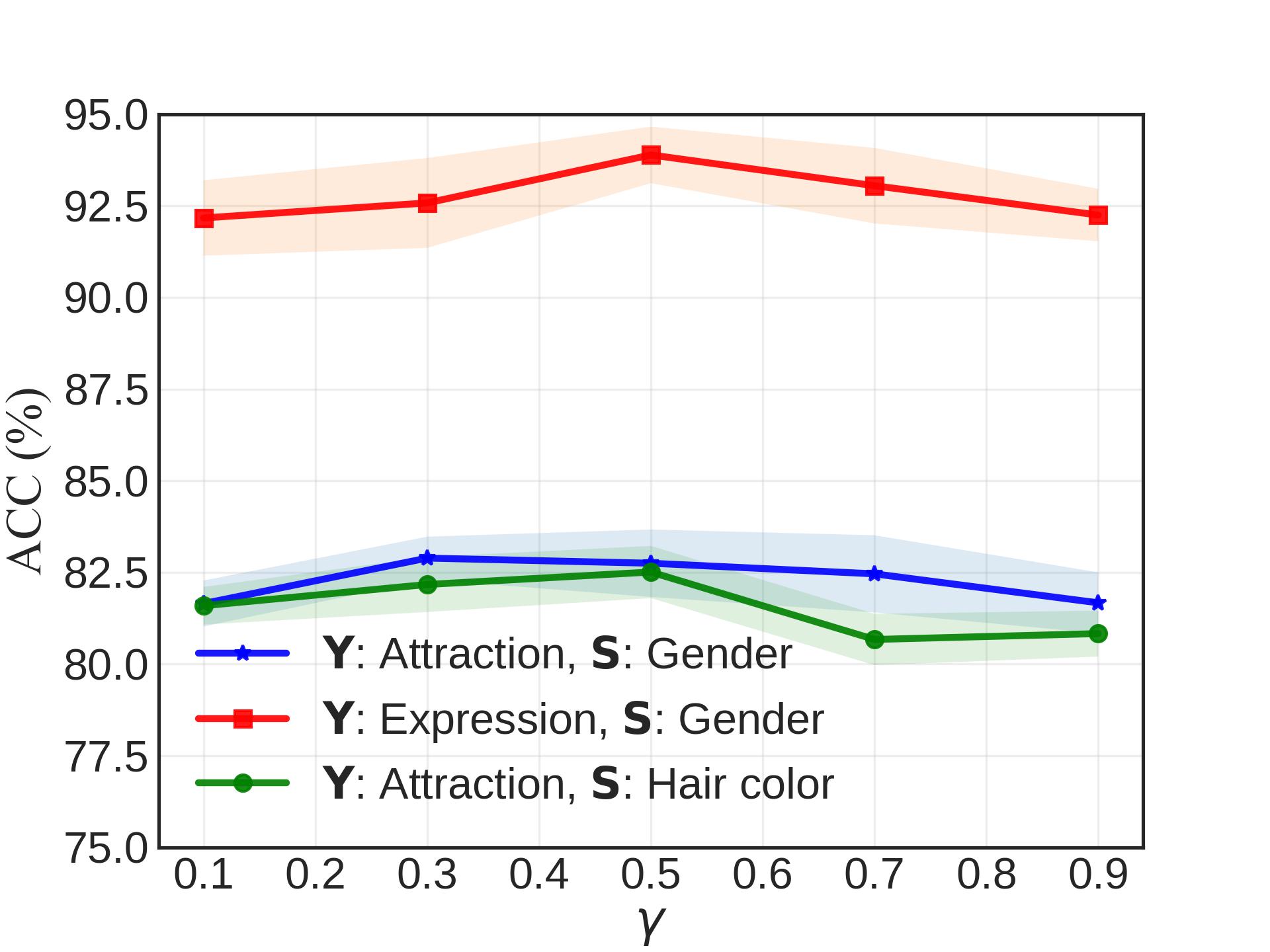}\label{fig:subfig7}}
  \subfloat
  {\includegraphics[width=0.49\textwidth]{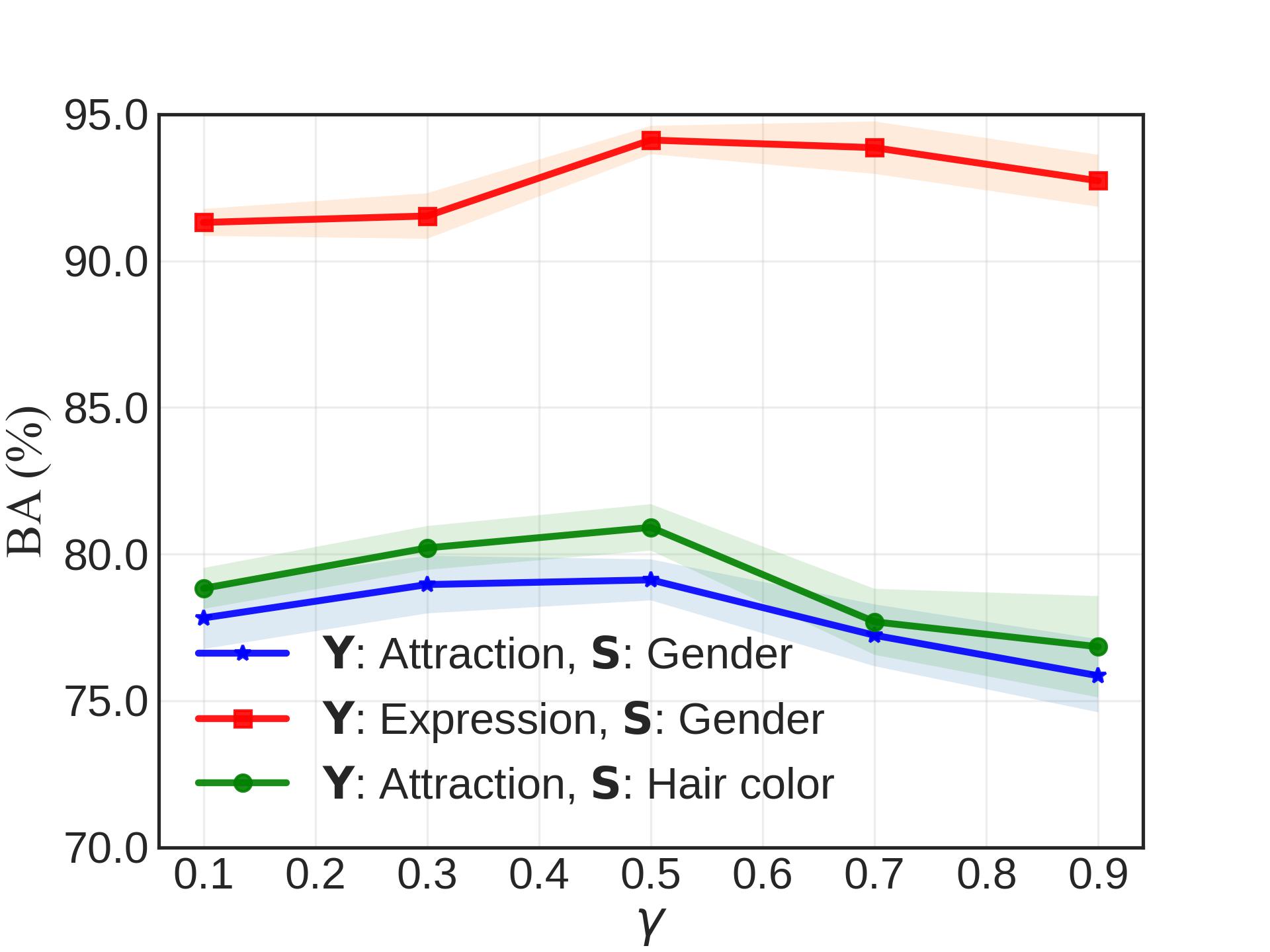}\label{fig:subfig8}}
\quad
  \subfloat
  {\includegraphics[width=0.49\textwidth]{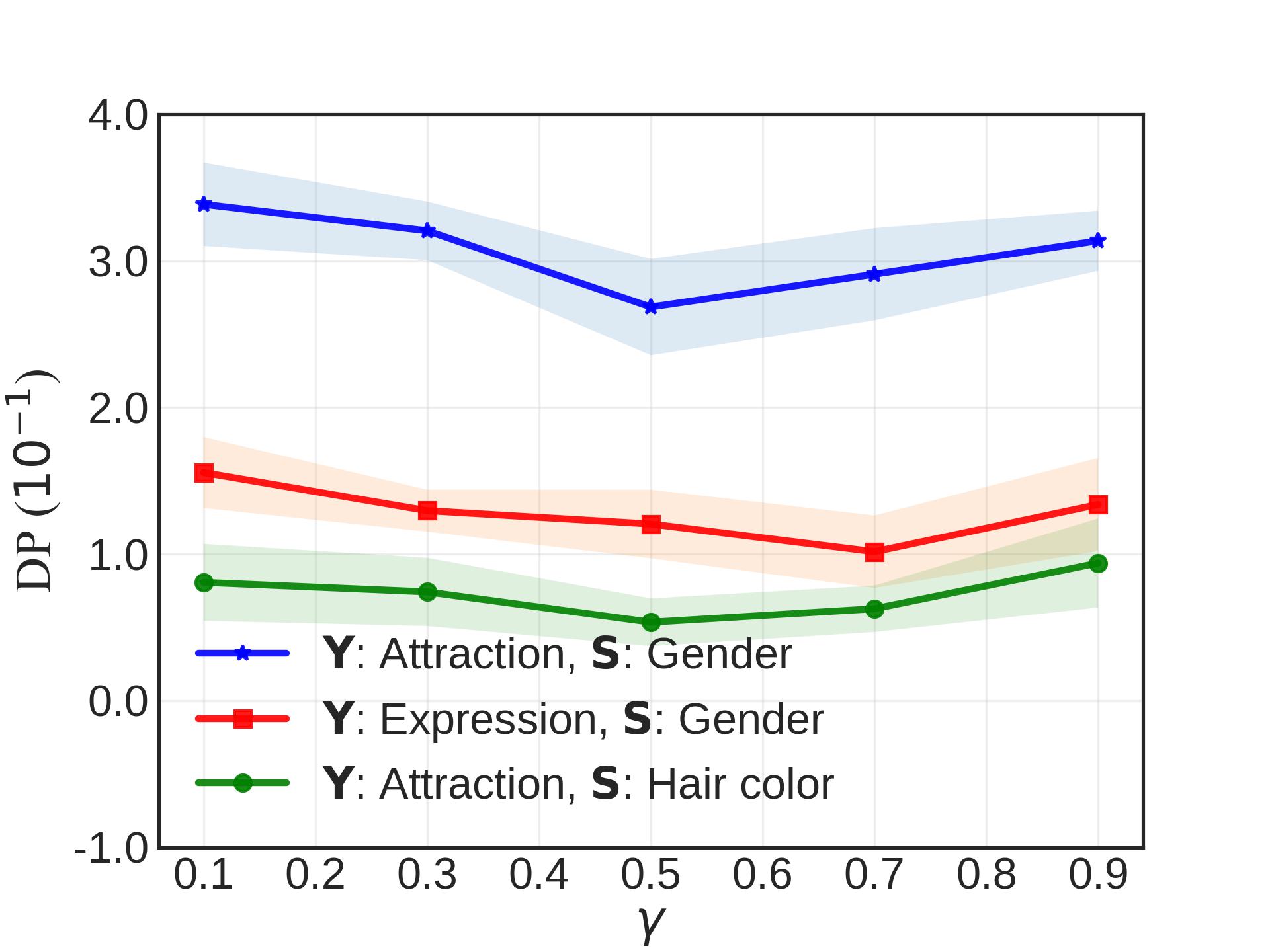}\label{fig:subfig9}}
  \subfloat
  {\includegraphics[width=0.49\textwidth]{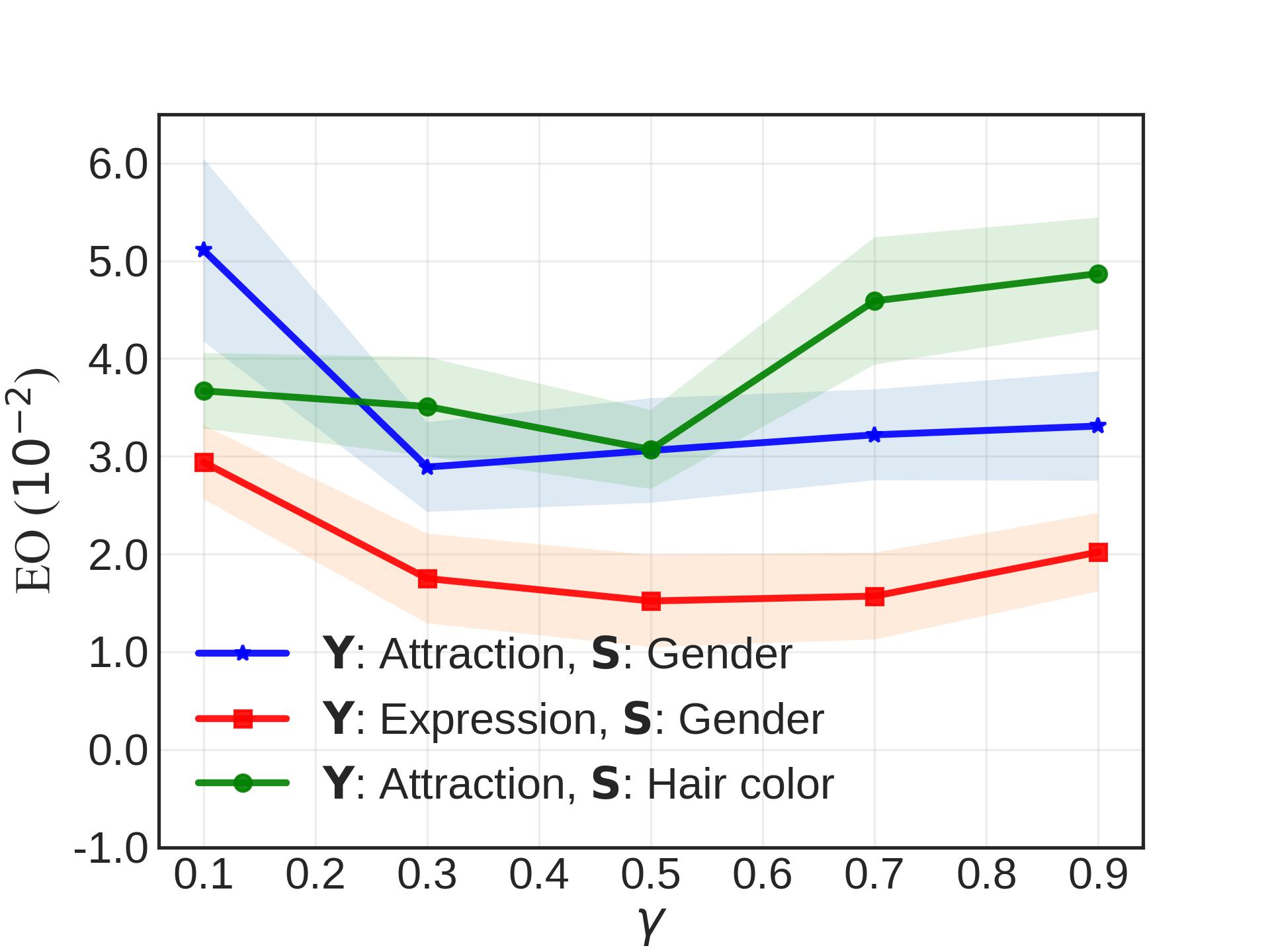}\label{fig:subfig10}}
  \caption{Impact of $\gamma$. Shown is the mean $\pm$ standard deviation of 3 runs with different random seeds.}
  \label{fig:gamma}
\end{figure}

Due to the limited space in Table \ref{tab:alpha} and \ref{tab:gamma}, we present the impact of $\alpha$ and $\gamma$ in a figure manner, adding the standard deviation to elaborate the systematic error in our experiments. The results are shown in Figure \ref{fig:alpha} and Figure \ref{fig:gamma}. 

\textbf{Impact of $\alpha$. }As $\alpha$ surpasses the threshold of $1$, a noticeable decline in accuracy is observed, coupled with an escalation in standard deviation. This suggests that a performance decline leads to a more fluctuating demonstration. Additionally, it is noteworthy that the results are not substantially sensitive to parameter selection for a reasonable range.

\textbf{Impact of $\gamma$.} Compared to $\alpha$, $\gamma$ maintains a comparatively low standard deviation and demonstrates less fluctuation across different values. There is a subtle performance peak observed at around $\gamma=0.5$.

\section{Implementation Details}
\label{sec:implementation}
The models are trained offline using PyTorch \cite{bib2} and executed on a machine equipped with an AMD Ryzen Threadripper 3970X 32-Core CPU @ 2.00GHz and an NVIDIA GeForce RTX A4000 GPU, running the Ubuntu 20.04 operating system. To ensure a consistent data flow during training and to save computing power, we opt to use the first 80 individuals from the CelebA dataset \cite{bib3} rather than the entire dataset.

To mitigate overfitting in the distance loss, we establish a lower bound of $-2$ for the calculation of each sample, further details are illustrated in the code. The code is available at \href{https://github.com/abdd68/Fair-Vision-Transformer}{https://github.com/abdd68/Fair-Vision-Transformer}.

\end{appendix}
\end{document}